\documentclass[preprint,12pt]{elsarticle}

\usepackage{amssymb}

\usepackage{booktabs}
\usepackage{subfigure}
\usepackage{subcaption} 

\usepackage{chngcntr}
\counterwithin{table}{section}

\usepackage{color}
\definecolor{red}{rgb}{0,0,0}
\definecolor{blue}{rgb}{0,0,1}
\usepackage{ulem}

\makeatletter
\renewcommand{\@thesubfigure}{\hskip\subfiglabelskip}
\makeatother

\begin{document}

\begin{frontmatter}



\title{PolyR-CNN: R-CNN for end-to-end polygonal building outline extraction}


\author[inst1]{Weiqin Jiao\corref{cor1}}

\affiliation[inst1]{organization={Department of Earth Observation Science, ITC Faculty Geo-Information Science and Earth Observation, University of Twente},
            addressline={Hallenweg 8}, 
            city={Enschede},
            postcode={7522 NH}, 
            state={Overijssel},
            country={The Netherlands}}
\cortext[cor1]{Corresponding author. Email: w.jiao@utwente.nl}
\author[inst1]{Claudio Persello}
\author[inst1]{George Vosselman}


\begin{abstract}
Polygonal building outline extraction has been a research focus in recent years. Most existing methods have addressed this challenging task by decomposing it into several subtasks and employing carefully designed architectures. Despite their accuracy, such pipelines often introduce inefficiencies during training and inference. This paper presents an end-to-end framework, denoted as PolyR-CNN, which offers an efficient and fully integrated approach to predict vectorized building polygons and bounding boxes directly from remotely sensed images. Notably, PolyR-CNN leverages solely the features of the Region of Interest (RoI) for the prediction, thereby mitigating the necessity for complex designs. Furthermore, we propose a novel scheme with PolyR-CNN to extract detailed outline information from polygon vertex coordinates, termed vertex proposal feature, to \textcolor{red}{guide} the RoI features to predict more regular buildings. PolyR-CNN demonstrates the capacity to deal with buildings with holes through a simple post-processing method on the Inria dataset. Comprehensive experiments conducted on the CrowdAI dataset show that PolyR-CNN achieves competitive accuracy compared to state-of-the-art methods while significantly improving computational efficiency, i.e., achieving 79.2 Average Precision (AP), exhibiting a 15.9 AP gain and operating 2.5 times faster and four times lighter than the well-established end-to-end method PolyWorld. Replacing the backbone with a simple ResNet-50, PolyR-CNN maintains a 71.1 AP while running four times faster than PolyWorld. \textcolor{red}{The code is available at: https://github.com/HeinzJiao/PolyR-CNN.}

\end{abstract}



\begin{keyword}
polygonal building outline extraction \sep polygon R-CNN \sep fully end-to-end \sep vertex proposal feature \sep buildings with holes
\end{keyword}

\end{frontmatter}


\section{Introduction}
\label{sec:intro}
Polygonal building outline extraction refers to generating building polygons in the vector format from aerial and satellite images, which is of great importance in many practical applications, such as topographic mapping \cite{li2019topological}, cadastral mapping \cite{10282644}, disaster management \cite{lu2004change} and urban planning \cite{yeh1999urban}. Compared with the raster data format, the vector representation of object outlines in the form of polygons is able to provide desired topological information, which is favoured by many operations in Geographic Information Systems (GIS) \cite{hu2022polybuilding}.

Extracting accurate building outlines is a challenging task. A building polygon should precisely resemble the building outline and avoid inaccurate shapes, such as rounded corners as a prediction for right angles. To tackle this challenge, most existing methods decompose the difficult task into several smaller tasks through a multi-model architecture. This has been the cornerstone to success in building outline extraction models. Current methods fall into two broad categories, namely the segmentation-based methods \cite{girard2021polygonal, zorzi2021machine, li2021joint, xu2022accurate} and the end-to-end methods \cite{castrejon2017annotating, acuna2018efficient, hu2022polybuilding, zorzi2022polyworld}. The segmentation-based methods first generate building masks through a building (semantic) segmentation network and then utilize a well-designed vectorization method to generate vectorized building polygons. End-to-end methods, conversely, predict building polygons in vector format directly. Earlier works \cite{castrejon2017annotating, acuna2018efficient, li2019topological, zhao2021building} employed a combined approach of Recurrent Neural Networks (RNN) and Convolutional Neural Networks (CNN) to predict building polygon corners sequentially. Alternative methods \cite{liang2019convolutional, wei2021graph} leverage Graph Convolutional Networks (GCN) to take over the RNN part, enabling simultaneous prediction of all building corners to improve efficiency. In contrast to the conventional CNN-RNN and CNN-GCN schemes, \cite{zorzi2022polyworld} introduced a novel strategy involving the simultaneous prediction of all building corners on the image, coupled with the generation of a permutation matrix to connect them in order. Despite the elimination of a building segmentation model, these end-to-end methods rely either on complex network structures like RNN or GCN \cite{castrejon2017annotating, acuna2018efficient, li2019topological, zhao2021building, liang2019convolutional, wei2021graph}, or well-designed multi-model architectures containing several consecutive modules for different subtasks, such as vertex detection, vertex position refinement and
adjacency matrix prediction \cite{zorzi2022polyworld}.

These well-established methods achieve good performance in both segmentation accuracy and polygonization regularity. Despite their success, it is worth noting that the existing methods incur some common limitations: 1) CNN-RNN/GCN schemes and multi-model architectures present inefficiencies in both training and inference processes. 2) End-to-end methods cannot handle complex cases like buildings with holes.

\begin{figure}[htbp]
\centering
\includegraphics[width=\textwidth]{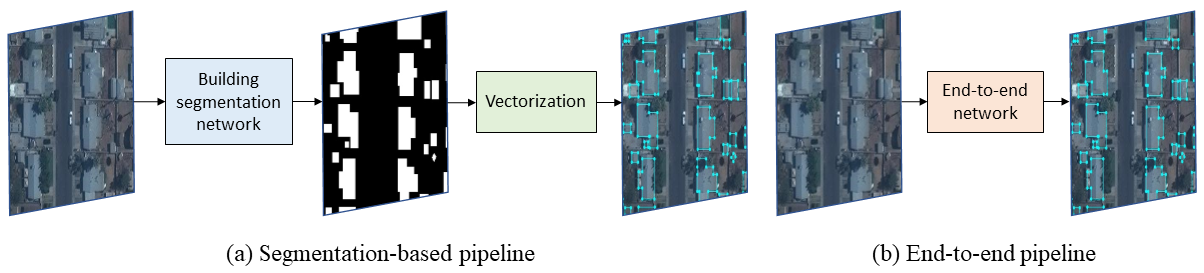}
\caption{\textcolor{red}{The two pipelines of existing polygonal building outline extraction methods.}}
\label{fig:comp}
\end{figure}

Although the multi-model architecture has been the mainstream of building outline extraction methods, a natural question arises: Is it possible to streamline the architecture and propose a simpler and more efficient method for building outline extraction? An intuitive idea is to view the building polygon as an extension of the bounding box in the object detection task and directly predict the vertex sequence of the building polygon in a correct connection order. Recently, PolyBuilding \cite{hu2022polybuilding} proposed to tailor a state-of-the-art object detection model, Deformable DETR \cite{zhu2020deformable}, to achieve this. Deformable DETR is built upon the Transformer architecture \cite{vaswani2017attention}, which has shown great performance in the field of natural language processing and computer vision. PolyBuilding extends it by adding two prediction heads, a polygon regression head and a corner classification head, parallel to the original prediction heads in Deformable DETR without additional modifications. The vertices of the reference polygons are uniformly sampled along the contours so that the polygon regression head can predict polygons with a fixed number of vertices. The corner classification head is used to differentiate valid corners from sampled points. The order of vertices reflects the adjacency between vertices. Despite its superior performance, PolyBuilding relies on the Transformer \cite{vaswani2017attention} architecture, which results in a long convergence time and low inference speed. In addition, the two prediction heads used to predict polygons directly use the features originally designed to predict object bounding boxes. These features often lack detailed instance characteristics, such as shape and pose, because the bounding box is only a rough representation of the instance geometry and location.

Given a small image area containing a building, humans can easily extract the outline of the building without seeing the whole image. Following this observation, the proposed method focuses on local areas inside the Region of Interest (RoI). The RoI features of CNN-based object detectors \cite{girshick2015fast, ren2015faster, girshick2014rich} are sufficient to predict accurate and regular building outlines. RoI features have already been utilized in previous methods \cite{li2019topological, zhao2021building}. However, these methods leverage RoI features for generating boundary and vertex masks, subsequently integrating them into the RNN module to predict building corners sequentially. In this paper, we propose a novel Region-based Convolutional Neural Network (R-CNN) named PolyR-CNN, which is conceptually simple, computationally efficient and fully end-to-end. PolyR-CNN directly predicts building polygons from RoI features, thereby obviating the need for complicated network structures. Inspired by recent set-based sparse object detectors \cite{sun2021sparse, carion2020end}, object candidates in PolyR-CNN are provided with a fixed small set of $N$ proposal bounding boxes and $N$ proposal polygons. The proposal polygon for each building instance is represented by a vertex sequence of a fixed length $M$. A corner classification head is utilized to distinguish corners from redundant vertices. Considering the fact that the RoI feature is only a coarse representation of the building, we also introduce a novel concept called \emph{vertex proposal feature} extracted from proposal polygons to encode detailed instance information into the RoI feature.

To evaluate the effectiveness of the proposed PolyR-CNN method, we conducted extensive experiments on the CrowdAI dataset \cite{mohanty2020deep} and the Inria dataset \cite{maggiori2017can}. Experimental results tested on the CrowdAI dataset demonstrate PolyR-CNN's outstanding performance. Compared with the CrowdAI dataset, the Inria dataset is more challenging. It contains buildings from different areas with different building distributions. It also contains a large amount of buildings with holes. To model these buildings, we consider cases where buildings have more than one polygon each (i.e., including polygons for one or multiple holes). To bypass this issue, PolyR-CNN views each valid polygon as an independent instance and uses a simple post-processing method to merge the predicted polygons belonging to the same building. Experimental results tested on the Inria dataset show that PolyR-CNN can generate outlines of buildings with holes, a characteristic not showcased by other end-to-end methods \cite{hu2022polybuilding, zorzi2022polyworld}. 

The main contributions of the paper are summarized as follows:
\begin{itemize}
\item We address polygonal building outline extraction through a streamlined deep learning architecture in a fully end-to-end manner \textcolor{red}{with superior computational efficiency}. This is also the first work to use an R-CNN model \textcolor{red}{to directly predict building polygons as ordered vertex sequences without any intermediate steps.}
\item We propose a scheme to exploit detailed instance characteristics from polygon coordinates to \textcolor{red}{guide} the RoI features to predict more regular buildings.
\item Experimental results on the CrowdAI dataset demonstrate that PolyR-CNN achieves comparable accuracy to the state-of-the-art methods, with shorter training convergence time, faster inference speed, and lower model complexity.
\end{itemize}

\section{Proposed Method}
\label{sec:m}
\begin{figure}[htbp]
\centering
\includegraphics[width=\textwidth]{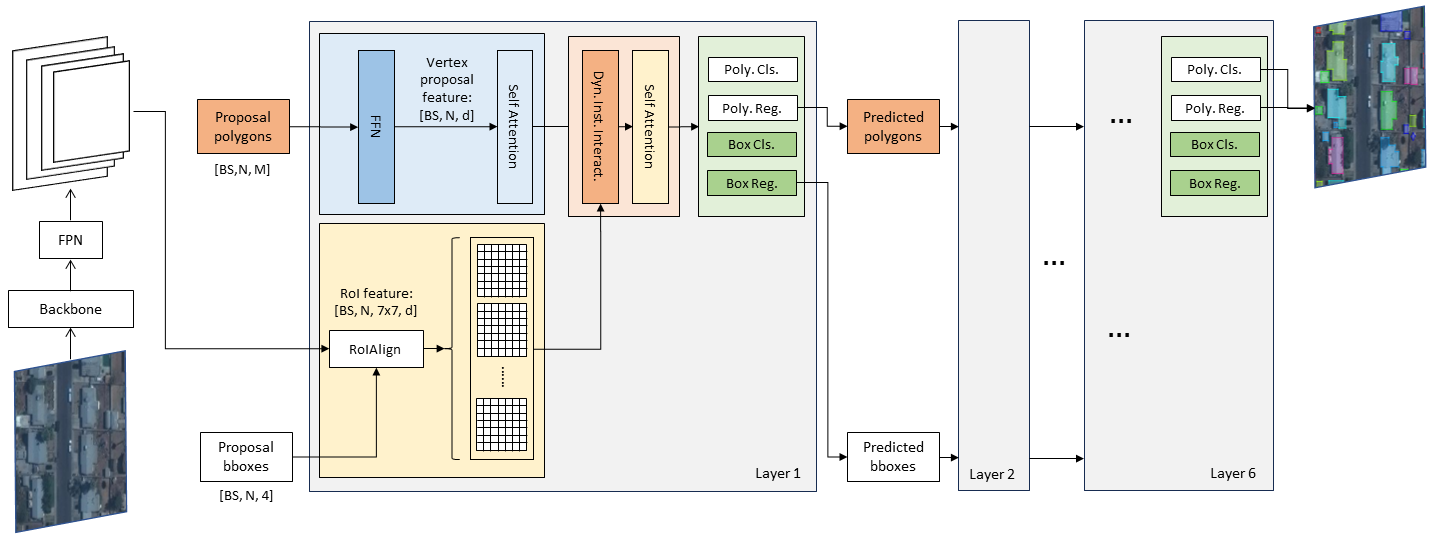}
\caption{The overall architecture of PolyR-CNN. The backbone provides multi-scale feature maps. The proposal polygons and proposal bounding boxes are initialized and iteratively refined through 6 consecutive layers. Each layer contains four modules: RoI feature extraction (light yellow), vertex proposal feature extraction (light blue), RoI feature \textcolor{red}{guidance} (light orange) and task-specific prediction heads (light green). The dimensions of each variable are also shown in the figure, where $BS$ denotes batch size, $N$ represents the number of proposals per image, $M$ is the unified number of vertices per polygon and $d$ is the feature dimension. The final building polygon is generated by filtering out redundant vertices based on the vertex classification scores.}
\label{fig:oa}
\end{figure}

The key idea of PolyR-CNN is to view building polygons as an extension of bounding boxes and utilize RoI features to generate the final predictions. Inspired by the success of SparseR-CNN \cite{sun2021sparse}, we follow its sparse design by initializing a small set of object proposals and applying set prediction loss \cite{carion2020end} on the predictions of bounding box classification scores, bounding box and polygon coordinates. The overall architecture is shown in Figure \ref{fig:oa}. PolyR-CNN consists of a backbone and 6 consecutive layers, each containing 3 main components: a vertex proposal feature extraction module, a RoI feature \textcolor{red}{guidance} module and four task-specific prediction heads. A detailed introduction of these components will follow in subsequent sections.

\subsection{Backbone}
\label{subsec:backbone}
ResNet \cite{he2016deep} architectures are employed as backbones in R-CNN models \cite{girshick2015fast, ren2015faster, girshick2014rich, sun2021sparse}. Following their scheme, we first adopt ResNet-50 \cite{he2016deep} and ResNeXt-101 \cite{xie2017aggregated} as the backbone to demonstrate the simplicity of our model. Given an input image $I\in\mathbb{R}^{3\times{H_0}\times{W_0}}$, a Feature Pyramid Network (FPN) \cite{lin2017feature} is used to extract multi-scale feature maps $P_l\in\mathbb{R}^{C\times{H_l}\times{W_l}}$ for $l \in \{2, 3, 4, 5\}$, where $H_l, W_l=\frac{H_0}{2^l}, \frac{W_0}{2^l}$ and $C=256$. Pre-trained weights on ImageNet-1K \cite{deng2009imagenet} are employed for initializing the ResNet backbones. The base Swin-Transformer (Swin-Base) \cite{liu2021swin} is also utilized as the backbone to provide multi-scale feature maps. The Swin-base backbone is initialized with weights pre-trained on ImageNet-21k \cite{deng2009imagenet}.

\subsection{Vertex proposal feature extraction}
Due to the ability of SparseR-CNN \cite{sun2021sparse} to effectively mitigate near-duplicates, we align the setting with it by using a fixed small number of proposal bounding boxes ($N\times4$) as region proposals to extract the RoI features. In addition to the proposal bounding boxes, we initialize a set of $N$ proposal polygons with a fixed number of $M$ vertices for each object instance. SparseR-CNN proved that the impact of initialization of proposal bounding boxes on the final results is negligible since their positions will undergo continuous refinement during training \cite{sun2021sparse}. Hence, following SparseR-CNN, we initialize the proposal bounding boxes using the 4-dimensional vector [0.5, 0.5, 1, 1], wherein the values represent normalized center coordinates, height, and width, respectively. The initial proposal polygons are generated by uniformly sampling $M$ vertices along the contour of the initial proposal bounding boxes. Since the bounding box covers not only the building itself but also its surroundings, the RoI feature extracted from the bounding box is only a coarse representation of the building. To remedy this shortcoming, we feed the proposal polygons ($N\times M$) into a feed-forward network (FFN) to generate a high-dimension vector ($C$) for each proposal polygon, which is also in one-to-one correspondence with the proposal bounding box and the RoI feature extracted based on the box. Vertex coordinates ($M\times2$) contain the geometric information of the building polygon, which can be learned through an FFN. We call such a high-dimensional vector a \emph{vertex proposal feature}, as it shares similarities with the concept termed \emph{proposal feature} in SparseR-CNN \cite{sun2021building}. The FFN contains two consecutive linear layers, followed by the Gaussian Error Linear Unit (GELU) \cite{hendrycks2016gaussian} activation function. Compared with the commonly used ReLU, GELU demonstrates higher nonlinearity and better performance in computer vision and natural language processing tasks. The extracted vertex proposal features are fed into a self-attention module \cite{vaswani2017attention} to enable global reasoning.With an iterative structure, the predicted bounding boxes and polygons will serve as the proposal bounding boxes and polygons of the next layer to offer more accurate geometrical information about the buildings.

\subsection{RoI feature \textcolor{red}{guidance}}
\label{subsec:roi feature guidance}
Similar to other R-CNN models \cite{girshick2015fast, ren2015faster, girshick2014rich}, we leverage RoIAlign \cite{he2017mask} to extract RoI features for each proposal bounding box. These RoI features are then enriched with instance characteristics through a Dynamic Instance Interactive Head \cite{sun2021sparse}, as introduced by SparseR-CNN \cite{sun2021sparse}. The interaction head contains two $1\times1$ convolutional layers, whose parameters are generated by the vertex proposal features through a linear layer. In this way, the RoI features passing through these two consecutive layers are regulated by the vertex proposal features. The resulting RoI features provide a comprehensive representation of individual buildings, encapsulating their distinctive attributes. However, they may lack the broader contextual information that could be useful in some cases, such as buildings close to each other. Therefore, we incorporate an additional self-attention module after the interactive head to let the RoI features interact and facilitate global reasoning capability among buildings. We note that self-attention modules are also utilized in PolyBuilding \cite{hu2022polybuilding}. However, the self-attention mechanism is the key to PolyBuilding's success, as the encoder-decoder Transformer architecture is built upon several successive self-attention modules.

\subsection{Task-specific prediction heads}
The object features output by the RoI feature \textcolor{red}{guidance} module are further fed into the task-specific prediction heads for the final predictions. In addition to the bounding box classification head and regression head inherited from object detection models, we add two heads, namely the polygon regression head and the vertex classification head. The former predicts $M$ vertices for each building polygon; the latter predicts a classification score for each vertex, based on which we filter out redundant vertices. For simplicity, we use a three-layer perceptron for each head. 

Due to the varying number of vertices in reference building polygons, we adopt the same uniform sampling strategy as PolyBuilding \cite{hu2022polybuilding} to unify the number of vertices to $M$ for the reference building polygons. We first sample $M$ vertices uniformly along the reference building contour and then replace the sampled vertex with its closest reference vertex since the $M$ uniformly sampled vertices may result in missing reference building vertices. We label the class of reference vertices as 1 and the sampled points as 0. The reference building polygons with an unified number of vertices and their vertex labels, after the aforementioned processing, will be utilized as the ground truth required for computing the loss.

\subsection{Set-based training losses}
Previous R-CNN methods \cite{girshick2015fast, ren2015faster, girshick2014rich} rely on densely-distributed object proposals. However, they resort to post-processing methods like Non-Maximum-Suppression (NMS) to eliminate near-duplicates caused by dense priors. To remove the reliance on a human-crafted process, we follow SparseR-CNN \cite{sun2021sparse} by adopting several proposal boxes and set-based losses \cite{carion2020end}. It first utilizes the Hungarian algorithm to perform bipartite matching between references and predictions and then calculates losses between the well-matched pairs. Given that the instance detection part of our task is relatively simple because it involves only one object class, we use only bounding box classification and regression losses as the matching cost. The matching cost $C$ is formally specified as follows:
\begin{equation}\label{equ:1}
	C = \lambda_{box}\cdot\mathcal{L}_{box}+\lambda_{cls}\cdot\mathcal{L}_{cls}+\lambda_{giou}\cdot\mathcal{L}_{giou}
\end{equation}
where $L_{box}$ and $L_{giou}$ are the L1 loss and the generalized IoU loss \cite{rezatofighi2019generalized} between the reference and predicted bounding boxes, $L_{cls}$ is the focal loss \cite{lin2017focal} for object classification, and $\lambda_{box}, \lambda_{giou}, \lambda_{cls}$ are the corresponding coefficients, which are hyperparameters.

The training losses consider both the bounding box and polygon:
\begin{equation}\label{equ:2}
	\mathcal{L} = \lambda_{box}\cdot\mathcal{L}_{box}+\lambda_{cls}\cdot\mathcal{L}_{cls}+\lambda_{giou}\cdot\mathcal{L}_{giou} + \lambda_{poly}\cdot\mathcal{L}_{poly} + \lambda_{vtx}\cdot\mathcal{L}_{vtx}
\end{equation}
where the first three terms are the same as the three of the matching cost, $\mathcal{L}_{poly}$ is the L1 loss between reference and predicted polygons, and $\mathcal{L}_{vtx}$ is the focal loss for polygon vertex classification.

\section{Experimental settings and dataset}
\subsection{Datasets}
We evaluate the performance of our proposed methods using two datasets related to vectorized builidng outline extraction, namely the CrowdAI Mapping Challenge dataset \cite{mohanty2020deep} (CrowdAI dataset) and the Inria Aerial Image Labeling dataset \cite{maggiori2017can} (Inria dataset). The CrowdAI dataset contains 280741 images for training and 60317 images for testing. \textcolor{red}{The size of each image is 300$\times$300 pixels} and the annotations are in MS-COCO \cite{lin2014microsoft} format. Compared with the CrowdAI dataset, the Inria dataset is more complex. It consists of 360 aerial images of \textcolor{red}{5000$\times$5000 pixels} from 10 different cities, whose building structures vary significantly. Moreover, a large proportion of these buildings exhibits internal holes, featuring multiple polygons per building. However, the original annotations of the Inria dataset are rasterized binary masks. Therefore, we converted them to vector format using the Douglas-Peucker method \cite{douglas1973algorithms, girard2021polygonal}. \textcolor{red}{Considering the complex shapes of building masks, we set the tolerance of the Douglas-Peucker algorithm to 5 pixels to simplify the obtained building polygons. We further merged polygon edges with angles smaller than $10^{\circ}$ or greater than $160^{\circ}$ to eliminate redundant structures derived from the masks.} As recommended by \cite{maggiori2017can}, we use the first five images of every city for validation. \textcolor{red}{Following \cite{xu2022accurate}, we crop the images to patches of 512$\times$512 pixels for training and validation.} The Inria dataset's test set contains the same number of images as the training and validation sets, originating from five distinct cities not encompassed in the training and validation sets. Since the Inria dataset's test set has not been released publicly, we conform to their official evaluation procedure. We convert the vectorized building outlines to segmentation masks and submit them to their official evaluation server.

\subsection{Evaluation metrics}
We evaluate the performance of PolyR-CNN based on three aspects: building instance segmentation, the quality of predicted building polygons and training and inference efficiency. To assess the building instance segmentation accuracy, we employ the standard MS-COCO \cite{lin2014microsoft} evaluation metrics, including Average Precision (AP) and Average Recall (AR), which are also the official evaluation criteria for the CrowdAI dataset. The computation of AP and AR involves the application of Intersection over Union (IoU) thresholds ranging from 0.50 to 0.95, with an interval of 0.5. We also compute $\rm AP_{50}$, $\rm AP_{75}$, $\rm AR_{50}$ and $\rm AR_{75}$, where the subscript denotes the IoU thresholds of 0.50 and 0.75, respectively. In addition, we also introduce the average boundary precision ($\rm AP_{boundary}$) and average boundary recall ($\rm AR_{boundary}$), which are calculated under a range of Boundary IoU (\cite{cheng2021boundary}) to give emphasis to the boundary quality. Given two masks $X$ and $Y$, the Boundary IoU first extracts the regions within a specified distance $d$ (0.02) from each contour and then calculates the IoU between them:
\begin{equation}
    {\rm Boundary \; IoU}{(X, Y)} = \frac{|(X_d\cap X)\cap(Y_d\cap Y)|}{|(X_d\cap X)\cup(Y_d\cap Y)|}
\end{equation}
The boundary IoU systematically measures the overlap degree between the contours of two masks, thereby quantifying the fidelity of predicted boundaries. 

Except for the mask IoU metrics, we also adopt the PoLiS metric \cite{avbelj2014metric} \textcolor{red}{and the N ratio \cite{zorzi2022polyworld} metric} to evaluate the quality of predicted polygons. Given two polygons $X$ and $Y$, the PoLiS distance is defined as follows:
\begin{equation}
    {\rm PoLiS}{(X, Y)} = \frac{1}{2m}\sum_{x_i \in X} {\mathop{\rm min}_{y\in\partial Y}||x_j - y||} + \frac{1}{2n}\sum_{y_j \in Y} {\mathop{\rm min}_{x\in\partial X}||y_j - x||}
\end{equation}
where the first term is the average distance between each vertex $x_i \in X, i = 1, ..., m$ of $X$ and its closest point $y$ on the boundary of polygon $Y$, while the second term is the average distances between each vertex $y_j \in Y, j = 1, ..., n$ of $Y$ and its closes point $x$ on the boundary of polygon $X$.

\textcolor{red}{The N ratio is a metric to measure the vertex redundancy of predicted polygons. It is the ratio between the vertex number of predicted and ground truth polygons. The closer the N ratio is to 1, the lower the vertex redundancy in the predicted polygon.}

We measure the training and inference efficiency of our model using three key metrics: total training epochs, Floating Point Operations (FLOPs) and Frames Per Second (FPS) during inference. Total training epochs refer to the number of epochs required to train the model until convergence. FLOPs serves as a measure of model complexity. It indicates the number of floating-point operations the model performs during inference. FPS reflects the inference speed. It represents the number of image frames the model can process per second during inference. In practical applications, it is often necessary to process large-scale aerial images. In such cases, low model complexity and fast model inference speed become particularly crucial.

\subsection{Experimental details}
We set the number of proposals per image $N$ to 100 and the unified vertex number per polygon $M$ to 96 for the CrowdAI dataset. \textcolor{red}{For the Inria dataset, $N$ is set to 300 and $M$ is set to 50.} These parameters are chosen to encompass the majority of reference building polygons in the datasets. We train our model on 2 Nvidia A40 GPUs with a batch size of 16. The initial learning rate is set to 2.5$\times 10^{-5}$. During training on the CrowdAI dataset, the models using ResNet-50 and Swin-Base as backbones are trained for 100 epochs. The learning rate decreases by a factor of 10 at epochs 54 and 83 for ResNet-50 and Swin-Base, respectively. For the ResNetXt-101 backbone, the training lasts 140 epochs, and the learning rate drops at epoch 74. For the Inria dataset, the model is trained for 100 epochs, utilizing Swin-base as the backbone, with a learning rate reduction at the $89^{th}$ epoch. We use only random flips as data augmentation for simplicity. Following \cite{hu2022polybuilding}, the coefficients in Equation \ref{equ:1} and \ref{equ:2} are set as $\lambda_{box}$=5, $\lambda_{cls}$=2, $\lambda_{giou}$=2, $\lambda_{poly}$=5 and $\lambda_{vtx}$=1. Like \cite{girard2021polygonal}, we keep polygons with more than 3 vertices and area larger than 10 pixels during evaluation.

\section{Experimental results}\label{sec:as}
\subsection{Comparison of state-of-the-art methods}
We compare the performance of PolyR-CNN with the state-of-the-art methods on the CrowdAI dataset, including 3 multi-stage and 3 end-to-end methods. To our knowledge, HiSup \cite{xu2022accurate} and PolyBuilding \cite{hu2022polybuilding} represent the most accurate state-of-the-art methods within the two respective categories. 

As shown in Table \ref{tab:1}, PolyR-CNN demonstrates comparable performance to state-of-the-art methods in building instance segmentation and polygon quality. Specifically, PolyR-CNN surpasses all end-to-end methods in MS-COCO metrics and $\rm AP_{boundary}$ by a large margin. Moreover, PolyR-CNN exhibits significant improvements in $\rm AP_{50}$, $\rm AP_{75}$, $\rm AR$, $\rm AR_{50}$ and $\rm AR_{75}$ compared with HiSup. It is noteworthy that both PolyBuilding and HiSup incorporate strong feature-extracting methods, such as stacked Transformer encoder layers and full-level hierarchical supervision learning schemes. In contrast, PolyR-CNN utilizes a simpler Feature Pyramid Network (FPN) yet achieves performance on par with these sophisticated methods, thereby underscoring PolyR-CNN's simplicity and efficiency. \textcolor{red}{However, the N ratio of PolyR-CNN is relatively higher, which can be attributed to the coarseness of RoI features for precise point-level detection. A more detailed discussion can be found in Section \ref{Sec:limit}.}

\begin{table}
\centering
\caption{Comparison with the state-of-the-art methods on CrowdAI dataset. The upper part of the table includes 3 multi-stage methods, while the lower part contains 4 end-to-end methods. The best scores in each category are in bold.}
\label{tab:1}
\resizebox{\linewidth}{!}{
\begin{tabular}{ccccccccccccccc}
   \toprule
   Method & \textcolor{red}{backbone} & $\uparrow \rm AP$ & $\uparrow \rm AP_{50}$ & $\uparrow \rm AP_{75}$ & $\uparrow \rm AR$ & $\uparrow \rm AR_{50}$ & $\uparrow \rm AR_{75}$ & $\uparrow \rm IoU$ & $\uparrow \rm AR_{boundary}$ & $\downarrow \rm PoLiS$ & \textcolor{red}{$\rightarrow$1 $\rm N \,ratio$} \\
   \midrule
   Li et al. \cite{li2021joint} & \textcolor{red}{UResNet101} & 73.8 & 92 & 81.9 & 72.6 & 90.5 & 80.7 & - & - & - & \textcolor{red}{-} \\
   FFL \cite{girard2021polygonal} & \textcolor{red}{UResNet101} & 67.0 & 92.1 & 75.6 & 73.2 & 93.5 & 81.1 & 84.3 & 34.4 & 1.945 & \textcolor{red}{1.13} \\
   HiSup \cite{xu2022accurate} & \textcolor{red}{HRNetV2-W48} & \textbf{79.4} & 92.7 & 85.3 & 81.5 & 93.1 & 86.7 & \textbf{94.3} & \textbf{66.5} & \textbf{0.726} & \textcolor{red}{1.02} \\
   \midrule
   PolyMapper \cite{li2019topological} & \textcolor{red}{VGG16} & 55.7 & 86 & 65.1 & 62.1 & 88.6 & 71.4 & 77.6 & 22.6 & 2.215 & \textcolor{red}{-} \\
   PolyWorld \cite{zorzi2022polyworld} & \textcolor{red}{R2U-Net} & 63.3 & 88.6 & 70.5 & 75.4 & 93.5 & 83.1 & 91.2 & 50.0 & 0.962 & \textcolor{red}{0.93} \\
   PolyBuilding \cite{hu2022po
   lybuilding} & \textcolor{red}{ResNet-50} & 78.7 & 96.3 & 89.2 & 84.2 & 97.3 & 92.9 & 94.0 & - & - & \textcolor{red}{\textbf{0.99}} \\
   PolyR-CNN & \textcolor{red}{Swin-Base} & 79.2 & \textbf{97.4} & \textbf{90} & \textbf{85.2} & \textbf{98.1} & \textbf{93.5} & 91.6 & 63.3 & 1.204 & \textcolor{red}{1.66} \\
   \bottomrule
\end{tabular}}
\end{table}

We also compared the training convergence speed, model complexity, and inference speed of PolyR-CNN against other \textcolor{red}{end-to-end} methods. Model complexity and inference speed are measured by Floating-point operations per second (FLOPs) and Frame Per Second (FPS), respectively. Both metrics are tested on one NVIDIA TESLA V100 GPU with batch size 1. \textcolor{red}{The input size is 320$\times$320 because the image is upsampled to be divisible by 32 or 64 for compatibility with the backbone in all the methods.} As seen in \textcolor{red}{Table \ref{tab:2}}, PolyR-CNN needs only half the training epochs required by PolyBuilding and runs significantly faster than the other methods during inference. The FLOPs results reveal that, even with a Swin-Transformer backbone, PolyR-CNN maintains a significantly simpler architecture than PolyWorld.

\begin{table}
\centering
\caption{Comparison of computational cost and efficiency among end-to-end methods on CrowdAI dataset. Total epochs refer to the number of epochs required to train the model. FLOPs and FPS are measured during inference on one NVIDIA TESLA V100 GPU with batch size 1. The best scores are in bold.}
\label{tab:2}
\resizebox{\linewidth}{!}{
\begin{tabular}{cccccc}
   \toprule
   Method & Backbone & $\downarrow$ Total epochs & $\downarrow$ FLOPs(G) & $\uparrow$ FPS \\
   \midrule
   PolyWorld \cite{zorzi2022polyworld} & - & - & 181.23 & 8.39 \\
   PolyBuilding \cite{hu2022polybuilding} & \textcolor{red}{ResNet-50} & 200 & \textbf{21.45} & 14.30 \\
   PolyR-CNN & ResNet-50 & \textbf{100} & 21.91 & \textbf{32.69} \\
   PolyR-CNN & ResNetXt-101 & 120 & 29.48 & \textbf{29.25} \\
   PolyR-CNN & Swin-Base & \textbf{100} & 46.55 & \textbf{20.50} \\
   \bottomrule
\end{tabular}}
\end{table}

\begin{figure}[!htbp]
\subfigure{
\begin{minipage}[t]{0.24\linewidth}
\centering
\includegraphics[width=\linewidth]{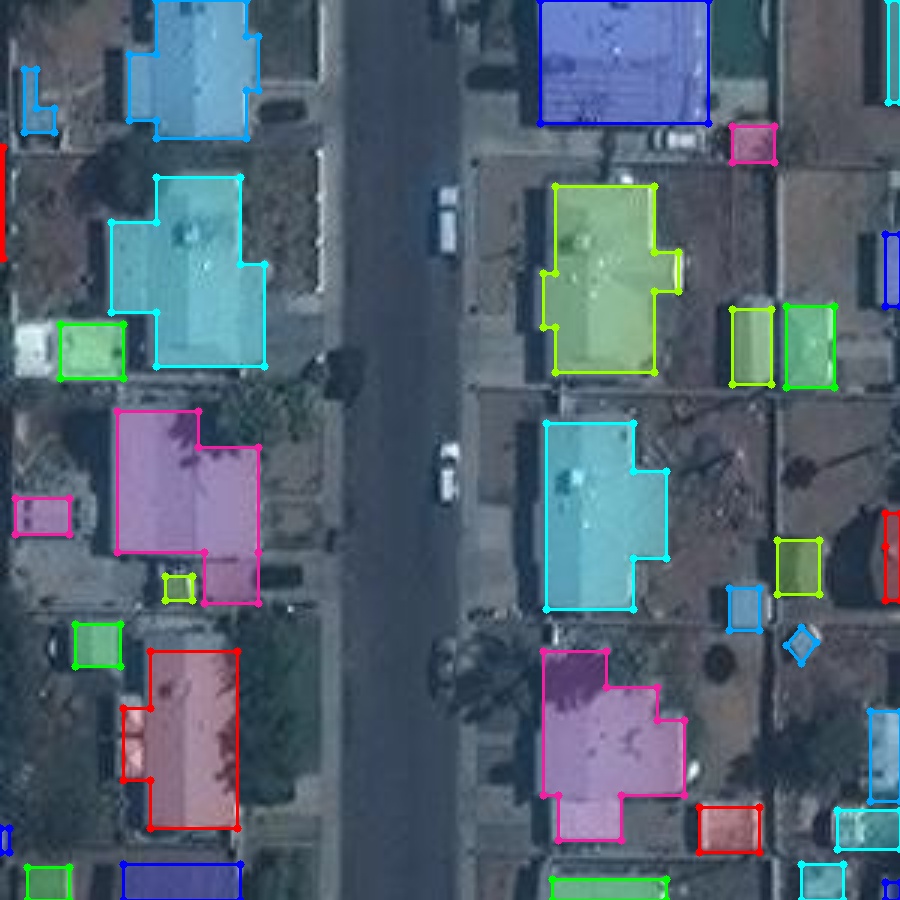}
\end{minipage}
}
\hspace{-4mm}
\subfigure{
\begin{minipage}[t]{0.24\linewidth}
\centering
\includegraphics[width=\linewidth]{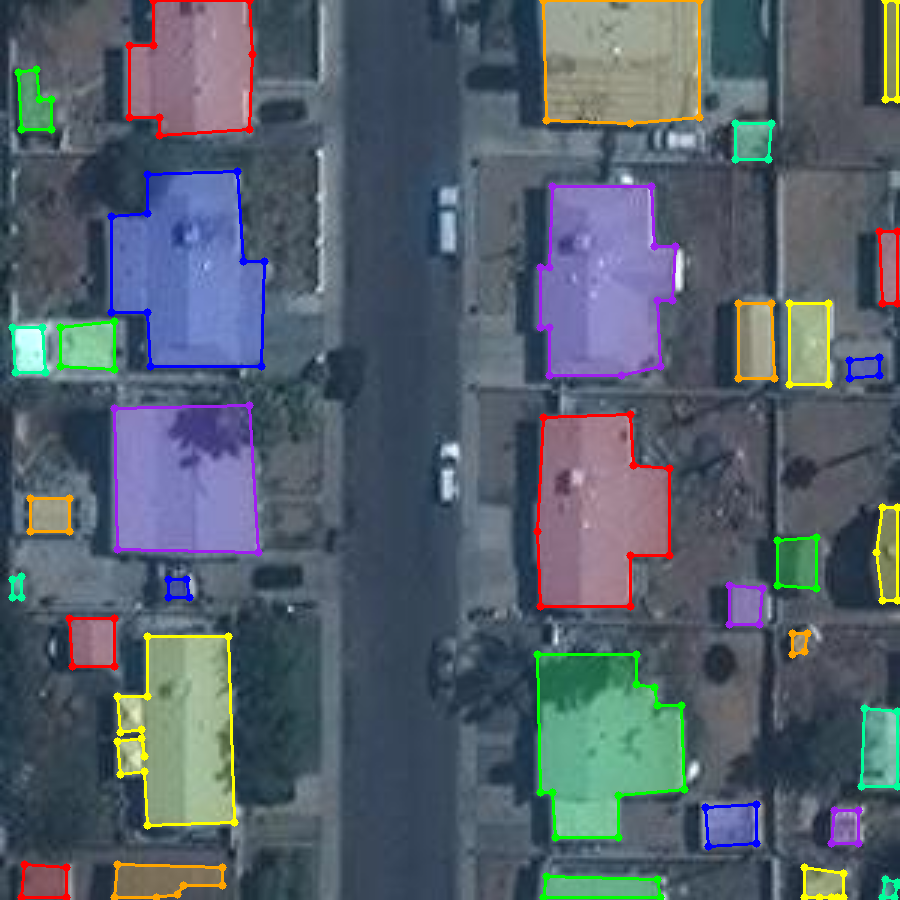}
\end{minipage}
}
\hspace{-4mm}
\subfigure{
\begin{minipage}[t]{0.24\linewidth}
\centering
\includegraphics[width=\linewidth]{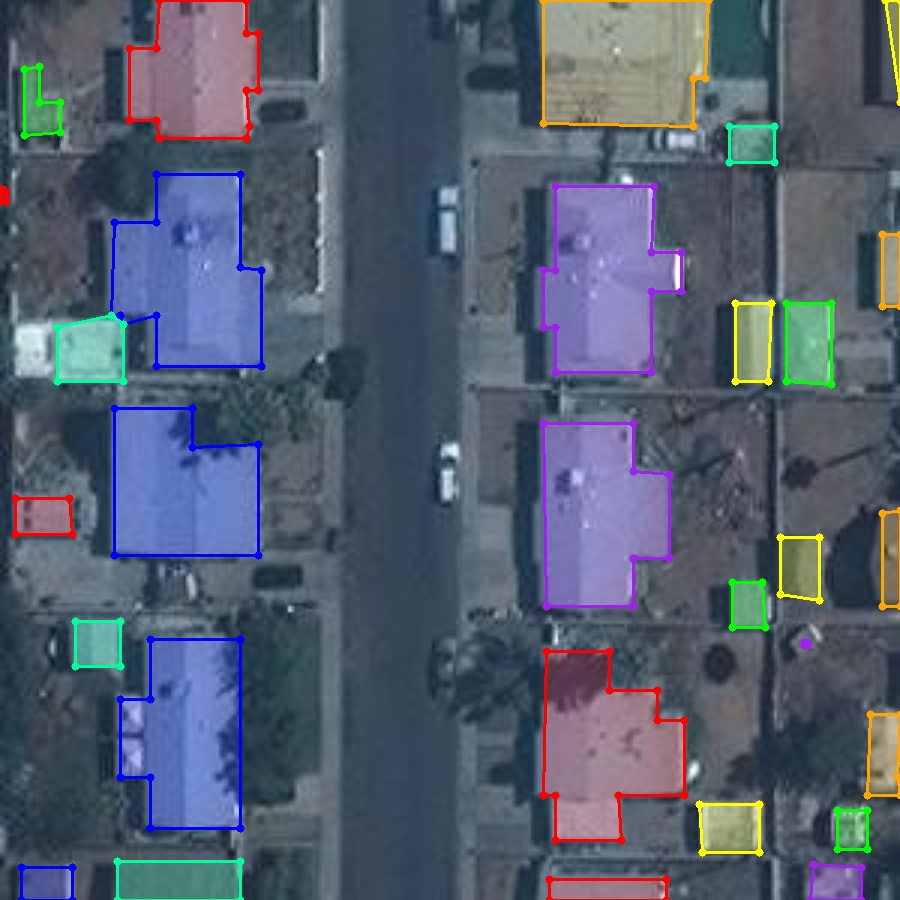}
\end{minipage}
}
\hspace{-4mm}
\subfigure{
\begin{minipage}[t]{0.24\linewidth}
\centering
\includegraphics[width=\linewidth]{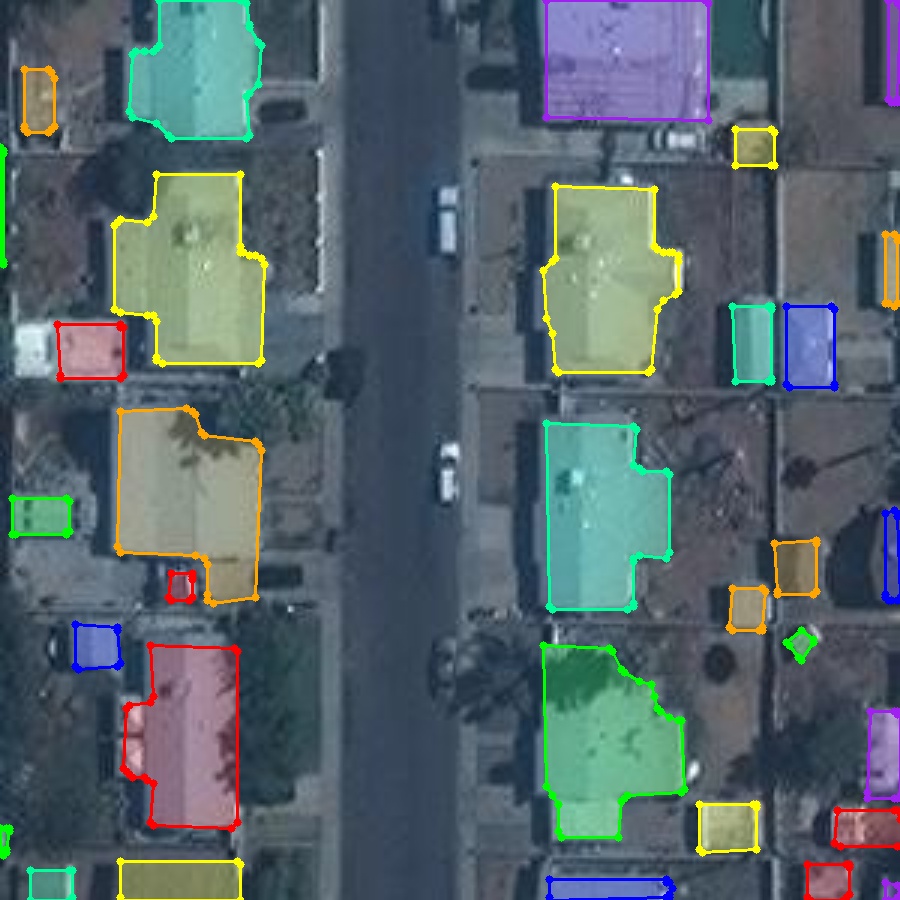}
\end{minipage}
}

\vspace{-3mm}

\subfigure{
\begin{minipage}[t]{0.24\linewidth}
\centering
\includegraphics[width=\linewidth]{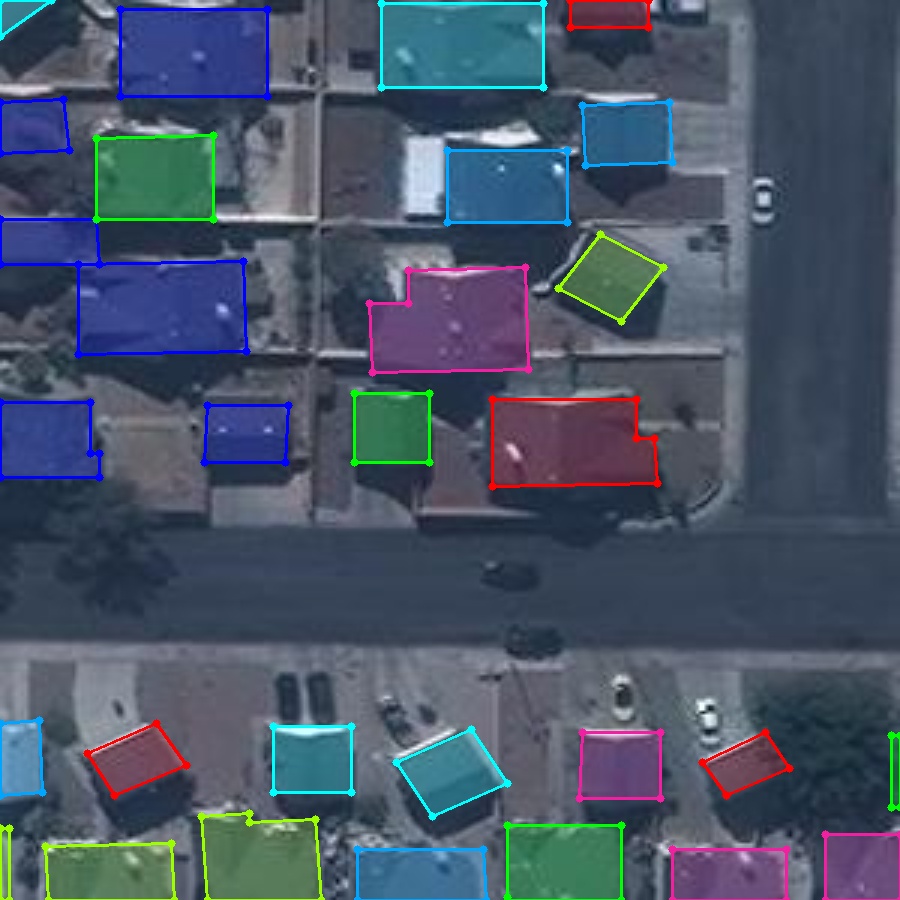}
\end{minipage}
}
\hspace{-4mm}
\subfigure{
\begin{minipage}[t]{0.24\linewidth}
\centering
\includegraphics[width=\linewidth]{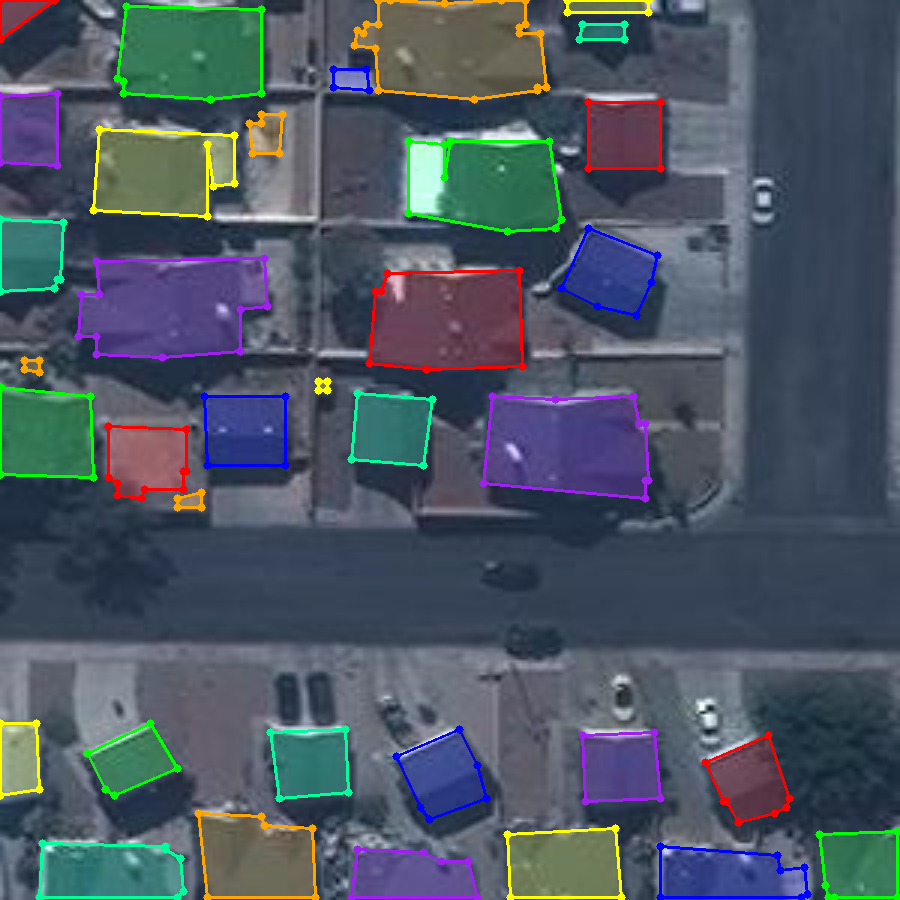}
\end{minipage}
}
\hspace{-4mm}
\subfigure{
\begin{minipage}[t]{0.24\linewidth}
\centering
\includegraphics[width=\linewidth]{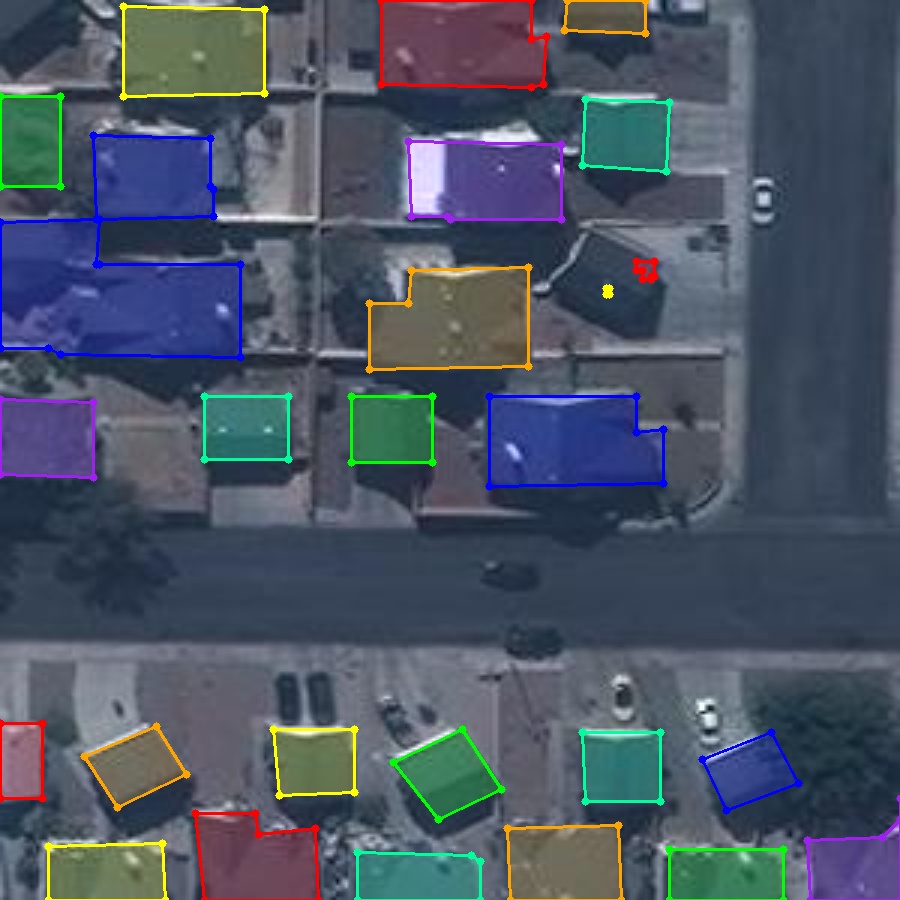}
\end{minipage}
}
\hspace{-4mm}
\subfigure{
\begin{minipage}[t]{0.24\linewidth}
\centering
\includegraphics[width=\linewidth]{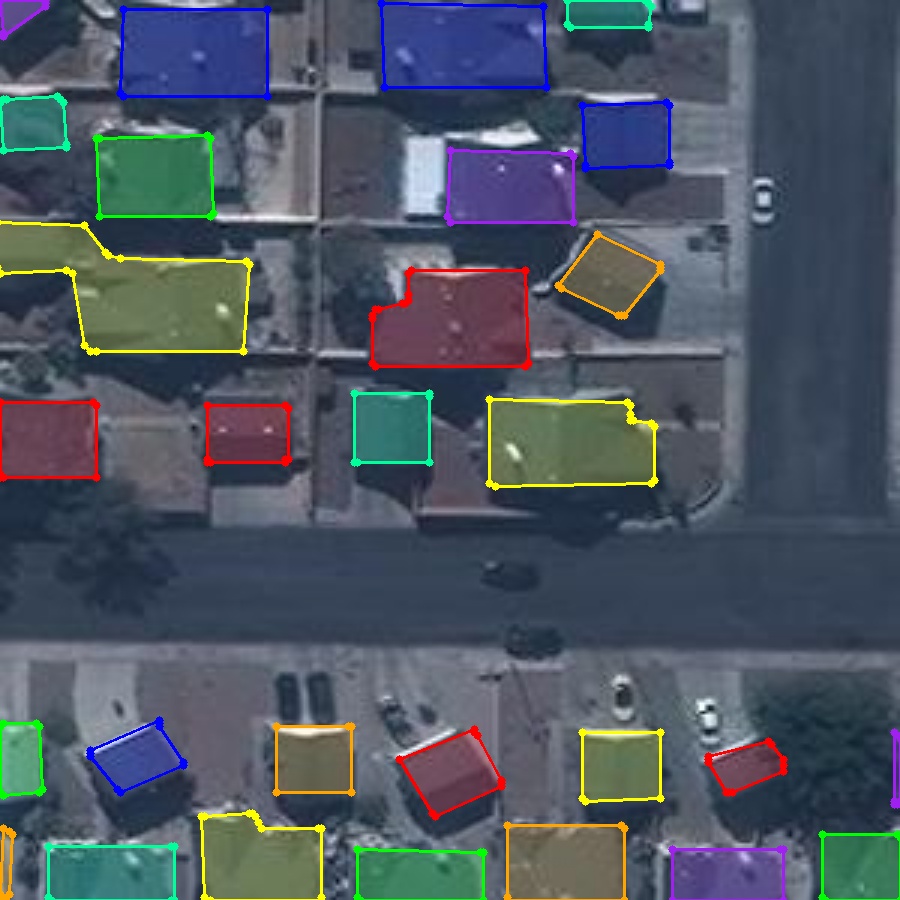}
\end{minipage}
}

\vspace{-3mm}

\subfigure{
\begin{minipage}[t]{0.24\linewidth}
\centering
\includegraphics[width=\linewidth]{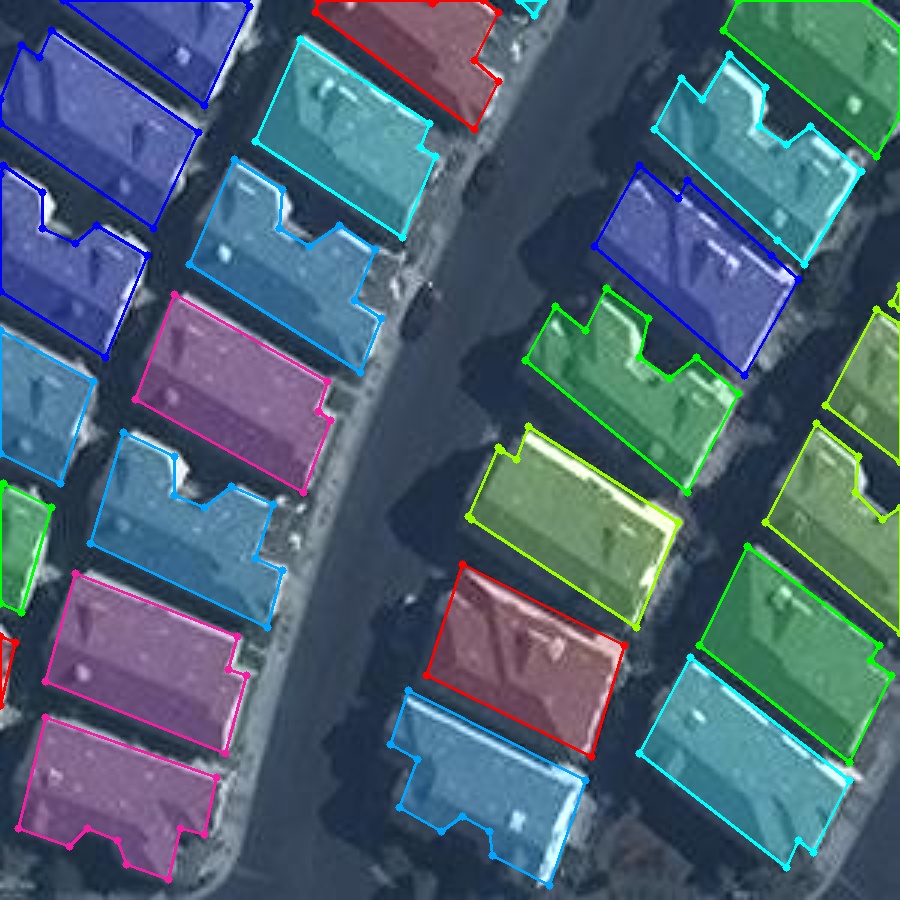}
\end{minipage}
}
\hspace{-4mm}
\subfigure{
\begin{minipage}[t]{0.24\linewidth}
\centering
\includegraphics[width=\linewidth]{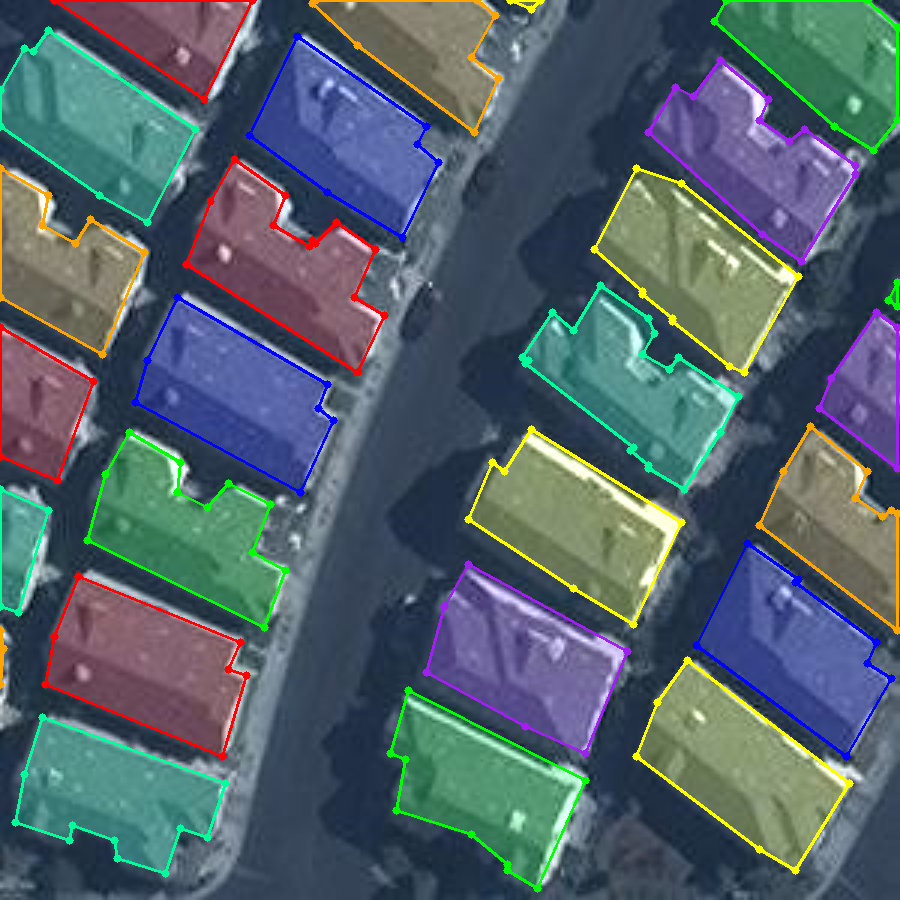}
\end{minipage}
}
\hspace{-4mm}
\subfigure{
\begin{minipage}[t]{0.24\linewidth}
\centering
\includegraphics[width=\linewidth]{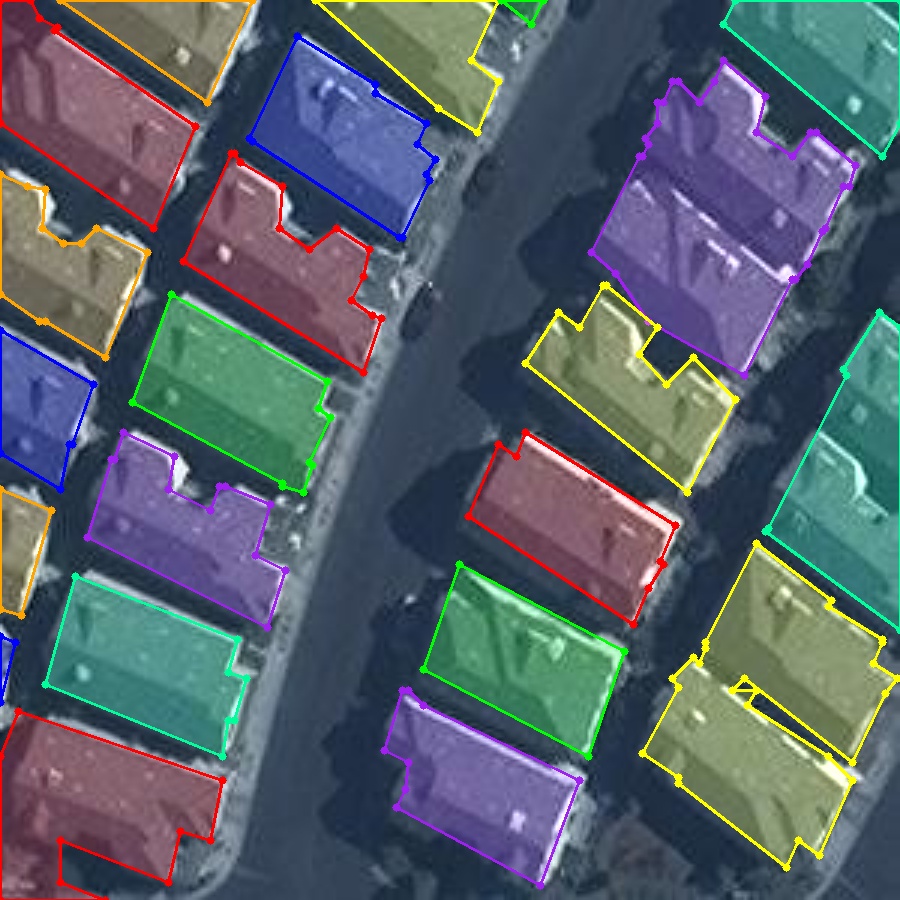}
\end{minipage}
}
\hspace{-4mm}
\subfigure{
\begin{minipage}[t]{0.24\linewidth}
\centering
\includegraphics[width=\linewidth]{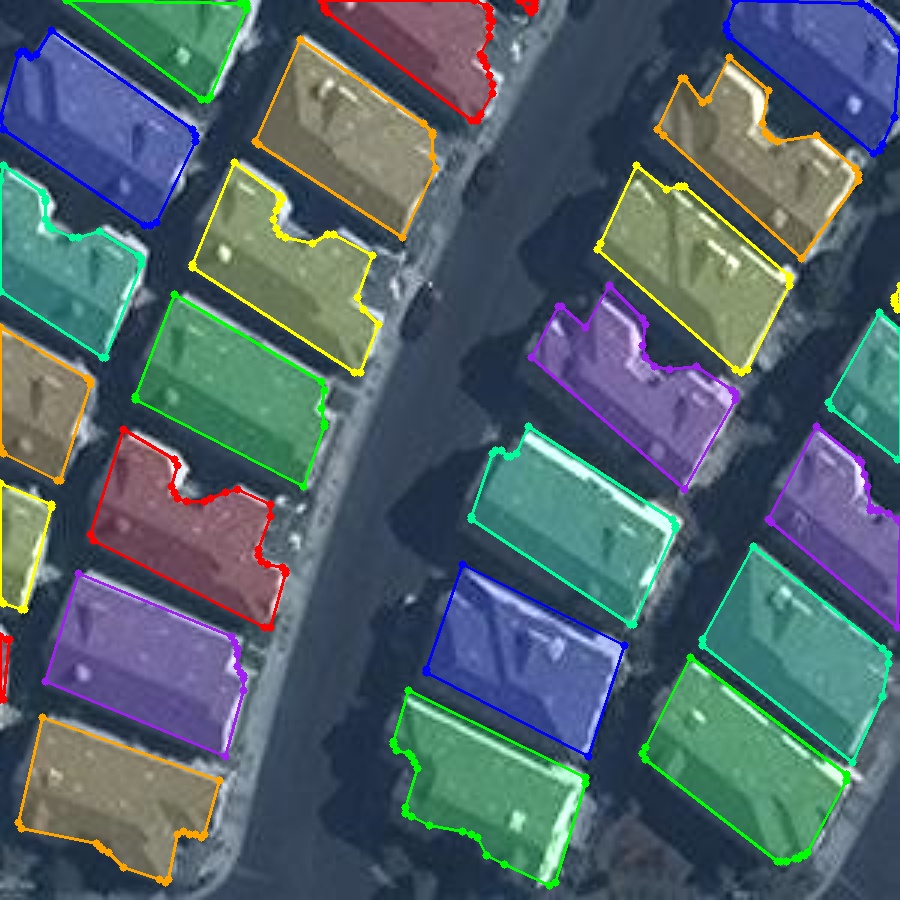}
\end{minipage}
}

\vspace{-3mm}

\subfigure[ground truth]{
\begin{minipage}[t]{0.24\linewidth}
\centering
\includegraphics[width=1\linewidth]{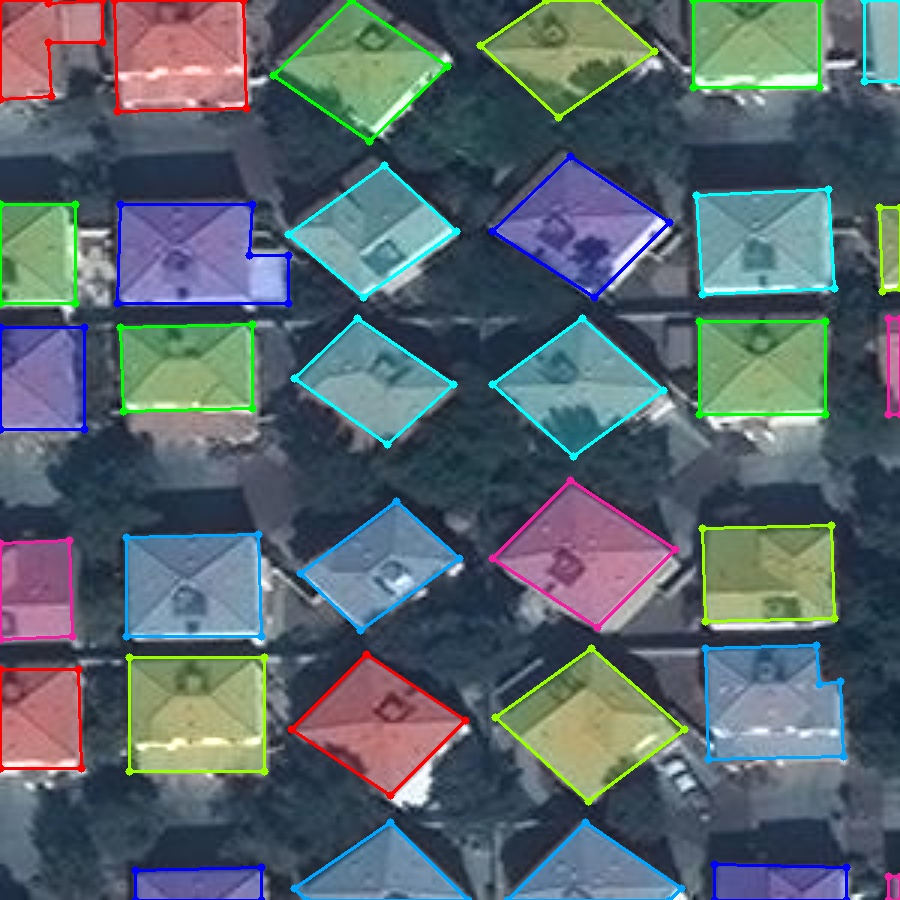}
\end{minipage}
}
\hspace{-4mm}
\subfigure[FFL]{
\begin{minipage}[t]{0.24\linewidth}
\centering
\includegraphics[width=1\linewidth]{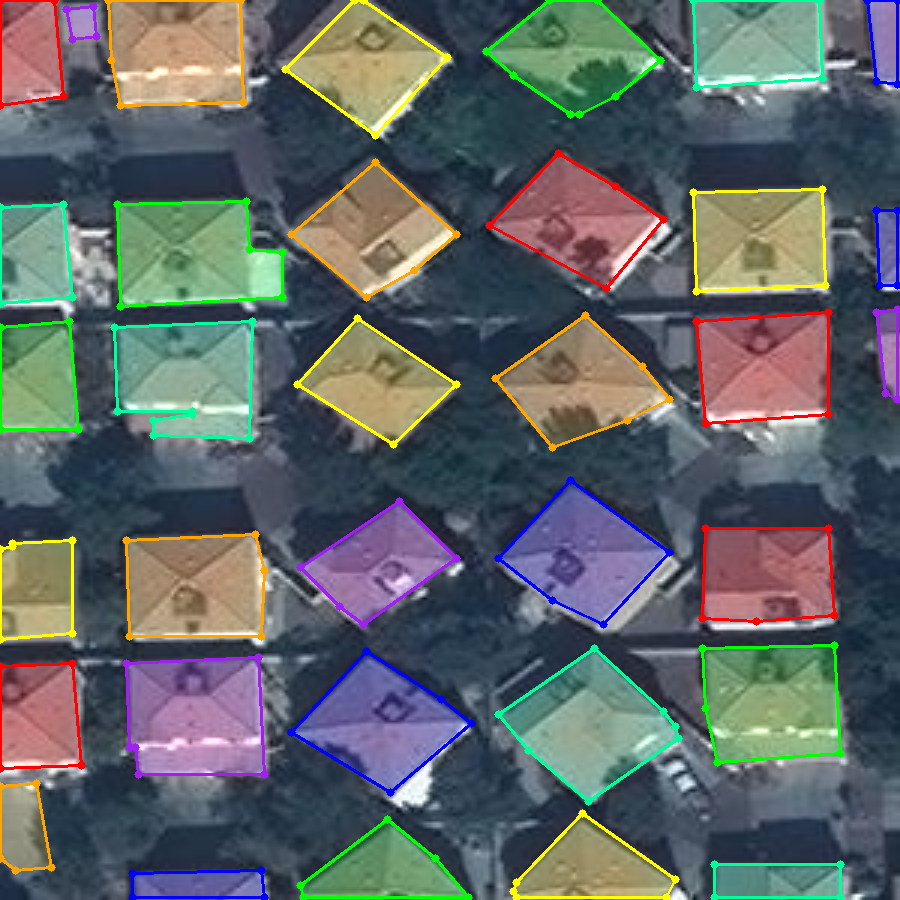}
\end{minipage}
}
\hspace{-4mm}
\subfigure[HiSup]{
\begin{minipage}[t]{0.24\linewidth}
\centering
\includegraphics[width=1\linewidth]{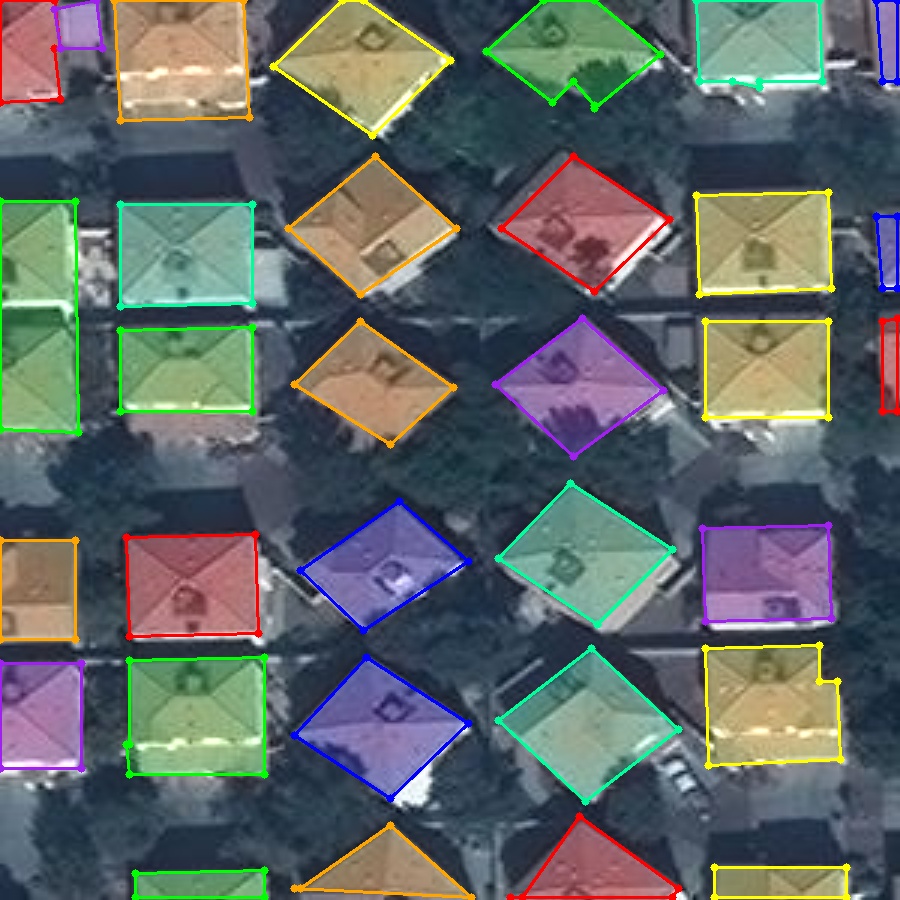}
\end{minipage}
}
\hspace{-4mm}
\subfigure[PolyR-CNN (ours)]{
\begin{minipage}[t]{0.24\linewidth}
\centering
\includegraphics[width=1\linewidth]{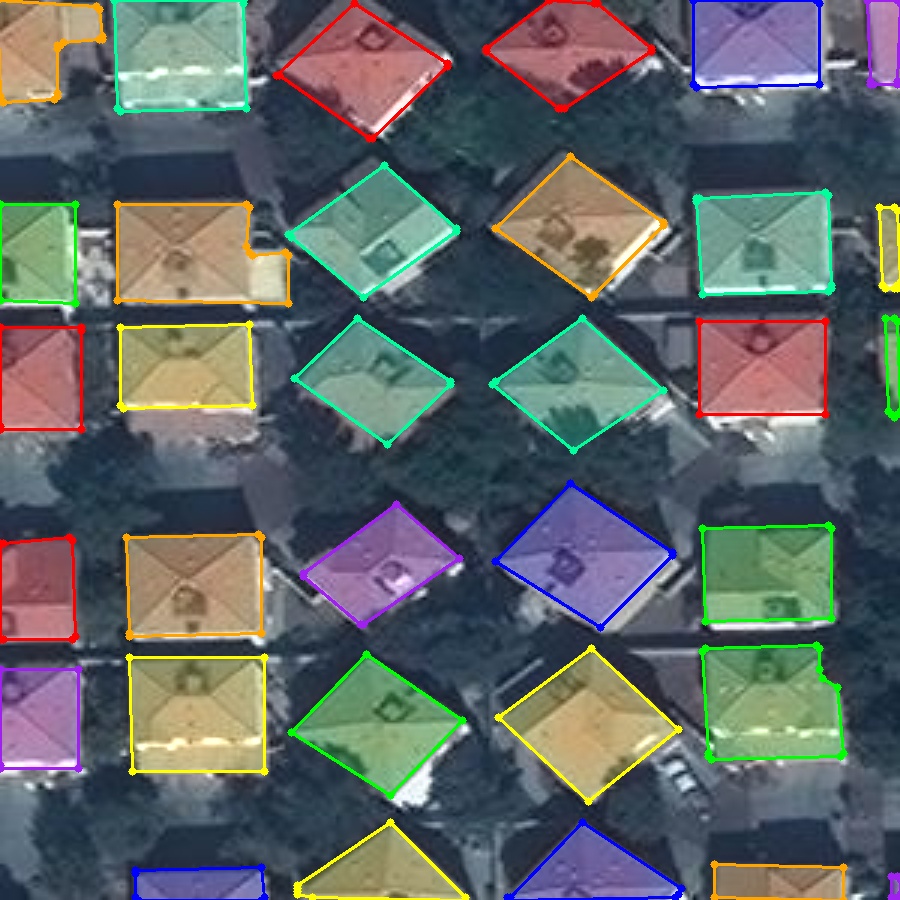}
\end{minipage}
}
\caption{Qualitative comparison of building extraction results on the CrowdAI test dataset among ground truth, FFL \cite{girard2021polygonal}, HiSup \cite{xu2022accurate} and PolyR-CNN (from left to right).}
\label{Fig:crowdai}
\end{figure}

Additionally, we present visualizations of some representative scenes to facilitate a qualitative comparison between PolyR-CNN, FFL, and HiSup. As shown in \textcolor{red}{Figure \ref{Fig:crowdai}}, PolyR-CNN demonstrates a high proficiency in detecting all buildings and accurately delineating their outlines. In comparison, although FFL can extract regular and accurate geometric outlines for certain buildings, it still exhibits a notable prevalence of false detections and is prone to generating intricate artefacts. HiSup demonstrates the capability to generate precise building outlines, even for buildings with complicated structures. However, when dealing with small buildings, HiSup is susceptible to miss detections and generation of degraded polygons. Moreover, when confronted with closely situated buildings, HiSup tends to produce overlapping edges and points and may erroneously treat several buildings as a single large structure. Last but not least, FFL and HiSup are prone to the influence of features resembling building outlines, such as rooftops and roads, leading to false detections. We hypothesize that this phenomenon arises due to the potential confusion induced by the frame field and attraction field map, regardless of whether it deals with small buildings, adjacent buildings, or features like rooftops and roads. In contrast, PolyR-CNN shows greater robustness in handling such scenarios as it does not rely on additional geometric supervision signals. 

\textcolor{red}{We further evaluate the performance of PolyR-CNN on the Inria dataset. Compared to the CrowdAI dataset, the Inria dataset presents greater challenges due to the diverse building types with complex geometries, especially for end-to-end methods lacking prior knowledge of building segmentation masks. 
Moreover, a large portion of the buildings includes inner polygons, further complicating the application of end-to-end methods on this dataset.
To address this issue, we treat each internal outline as an individual instance, considering them as a different class compared to the outer polygons, and subsequently group the predicted polygons of the same building.}
The performance measures provided by the official evaluation server focus on building semantic segmentation, including Intersection over Union (IoU) and Accuracy (the percentage of correctly classified pixels). 
\textcolor{red}{To demonstrate the computational efficiency of our method, we also report the inference speed measured by Frame Per Second (FPS). 
The FPS metrics for PolyR-CNN and HiSup are tested on a NVIDIA GTX 3090Ti GPU with a batch size of 1. The input size is 512$\times$512.} 

\textcolor{red}{As the first end-to-end method undergoing the official training and testing procedure on the Inria dataset, we compare our method with two segmentation-based polygonal building delineation methods, HiSup \cite{xu2022accurate} and Zorzi et al. \cite{zorzi2019regularization}. 
Notably, PolyBuilding \cite{hu2022polybuilding} also provides experimental results on the Inria dataset, however, it does not follow the official evaluation protocol. Hence, we do not include PolyBuilding in the comparison. 
As shown in Table \ref{tab:6}, despite the absence of building masks as prior knowledge, our method still outperforms Zorzi et al. \cite{zorzi2019regularization}
on both IoU and Accuracy. 
Among polygonal building extraction methods, HiSup achieves the best performance on the Inria dataset. 
However, its reliance on subsequent polygonization methods results in lower inference speed compared to PolyR-CNN. 
This discrepancy will be particularly noteworthy when dealing with large-scale remote sensing data in practical scenarios.
The gap in IoU and accuracy is explainable considering that segmentation-based methods learn building masks directly, giving them an inherent advantage over end-to-end methods on semantic segmentation metrics.
Besides, the Inria dataset contains buildings with much more complex structures than the CrowdAI dataset, which poses increased challenges for PolyR-CNN to locate the corners directly.
}

\begin{figure}[!htbp]
\subfigure[HiSup]{
\begin{minipage}[t]{0.48\linewidth}
\centering
\includegraphics[width=\linewidth]{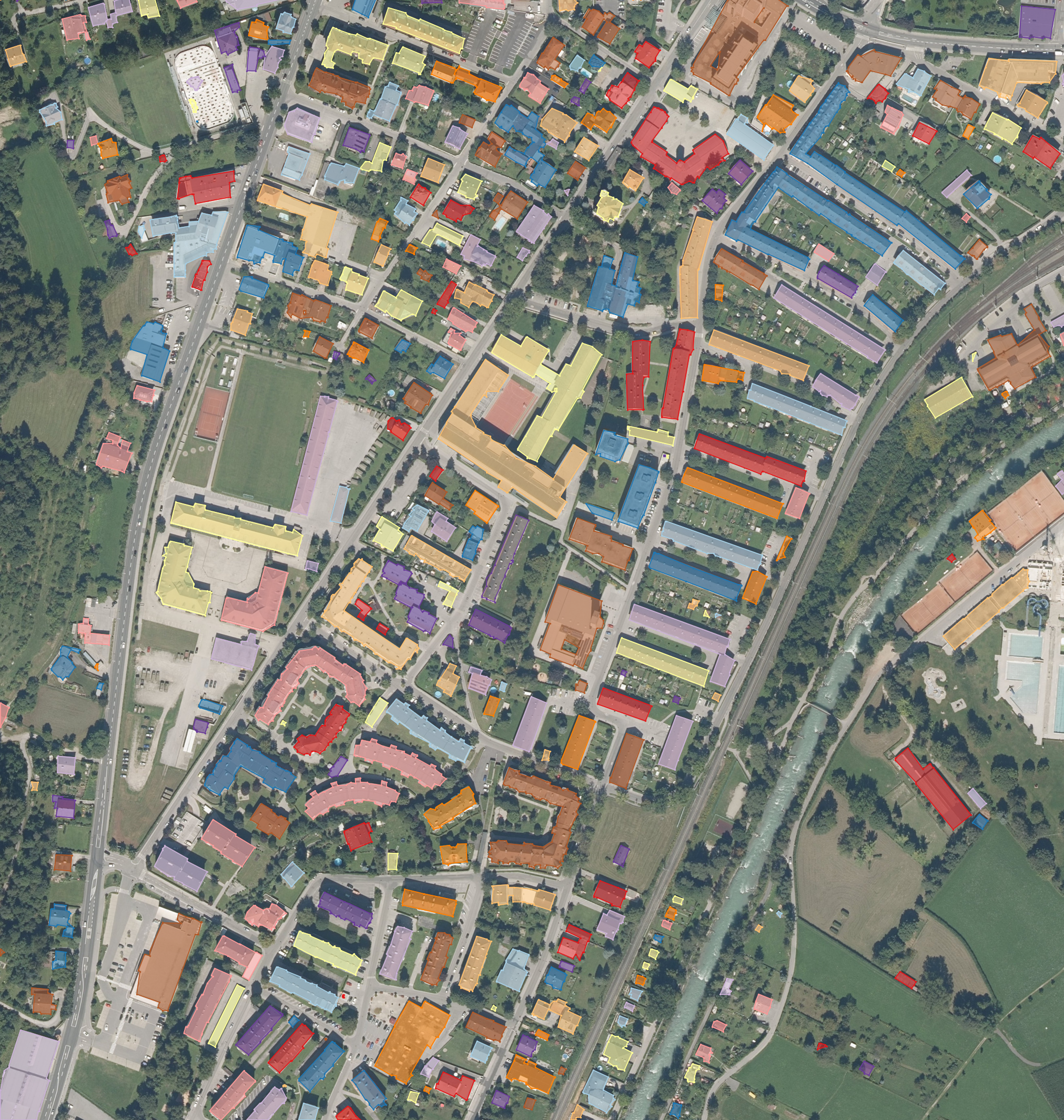}
\end{minipage}
}
\hspace{-4mm}
\subfigure[PolyR-CNN (ours)]{
\begin{minipage}[t]{0.48\linewidth}
\centering
\includegraphics[width=\linewidth]{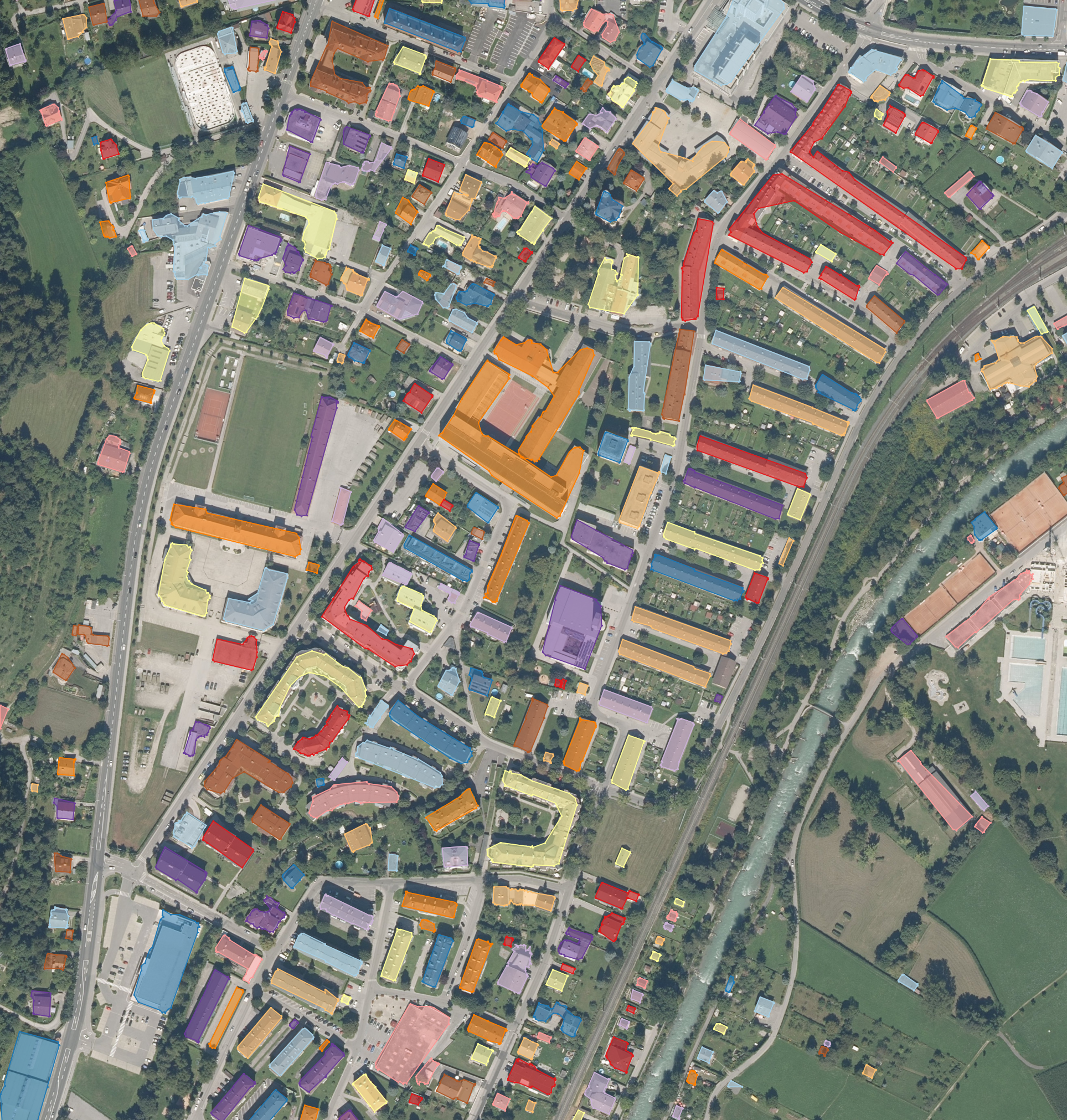}
\end{minipage}
}
\caption{\textcolor{red}{Qualitative comparison on the Inria test dataset between HiSup \cite{xu2022accurate} and PolyR-CNN.}}
\label{Fig:inria}
\end{figure}

\begin{figure}[!htbp]
\centering
\includegraphics[width=1\linewidth]{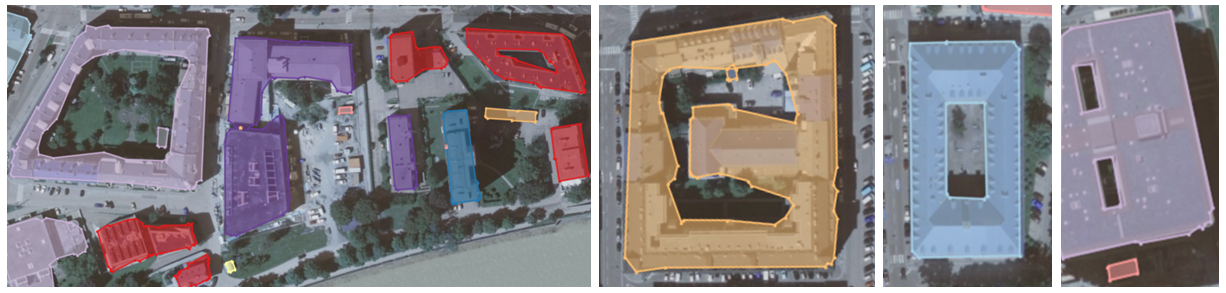}
\caption{Examples of predicted building outlines with internal holes of the Inria test images.}
\label{Fig:inner_hole}
\end{figure}

\textcolor{red}{Figure \ref{Fig:inria} shows the qualitative results of PolyR-CNN on a crop of a test image from the Inria dataset compared to HiSup. Despite not having building masks as prior knowledge, PolyR-CNN is still able to extract polygonal building outlines with a relatively high accuracy.} In Figure \ref{Fig:inner_hole}, we also visualize some examples to show the effectiveness of PolyR-CNN in extracting buildings with holes. However, PolyR-CNN exhibits a tendency to generate irregular shapes, especially for inner outlines. Addressing this limitation represents a key focus in our future work. 

\begin{table}
\centering
\caption{\textcolor{red}{Quantitative comparison with segmentation-based polygonal building extraction methods on the Inria dataset. The best scores are in bold.}}
\label{tab:6}
\begin{tabular}{ccccc}
   \toprule
   Method & segmentation required & $\uparrow$ IoU & $\uparrow$ Acc. & $\uparrow$ FPS \\
   \midrule
   Zorzi et al. \cite{zorzi2019regularization} & $\checkmark$ & 59.81 & 93.92 & - \\
   HiSup \cite{xu2022accurate} & $\checkmark$ & \textbf{75.53} & \textbf{96.27} & 16.41 \\
   PolyR-CNN & - & 68.35 & 95.09 & \textbf{35.73} \\
   \bottomrule
\end{tabular}
\end{table}

\subsection{Ablation study}
In this section, we analyze the effectiveness of the key components in PolyR-CNN, including the backbone, the self-attention module, and the vertex proposal feature.

\noindent\textbf{Influence of different backbones}. Due to the absence of strong feature extraction methods like those in PolyBuilding \cite{hu2022polybuilding}, and HiSup \cite{xu2022accurate}, the features provided by the backbone play an important role in PolyR-CNN. As mentioned in Section \ref{subsec:backbone}, three different pre-trained models were employed as backbones, namely ResNet-50, ResNetXt-101 and base Swin-Transformer. As can be seen in \textcolor{red}{Table \ref{tab:3}}, the base Swin-Transformer exhibits superior performance, followed by ResNetXt-101 and ResNet-50. Compared to ResNet-50, ResNetXt-101 has a deeper network structure, enabling the extraction of more salient information from images and, consequently, better performance in building outline extraction. However, the improvement is accompanied by a longer model convergence time. Different from the ResNet architectures, Swin-Transformer excels in extracting global information, facilitating the extraction of all building outlines in an image. Therefore, it achieves the best quantitative results, indicating that PolyR-CNN has the potential to take advantage of more advanced backbones like Deformable DETR \cite{zhu2020deformable}, upon which PolyBuilding \cite{hu2022polybuilding} is built. Nonetheless, the enhancement comes at the expense of increased model complexity and inference time, as shown in \textcolor{red}{Table \ref{tab:2}}.

\begin{table}
\centering
\caption{Ablation studies on different backbones in PolyR-CNN on CrowdAI dataset.}
\label{tab:3}
\resizebox{\linewidth}{!}{
\begin{tabular}{ccccccccccccccc}
   \toprule
   Backbone & Epoch & $\uparrow \rm AP$ & $\uparrow \rm AP_{50}$ & $\uparrow \rm AP_{75}$ & $\uparrow \rm AR$ & $\uparrow \rm AR_{50}$ & $\uparrow \rm AR_{75}$ & $\uparrow \rm AP_{boundary}$ & $\downarrow \rm PoLiS$ \\
   \midrule
   ResNet-50 & 100 & 71.1 & 93.8 & 82.9 & 78.6 & 95.6 & 88.3 & 50.0 & 1.570 \\
   ResNetXt-101 & 120 & 75.4 & 95.1 & 86 & 81.8 & 96.7 & 90.4 & 55.7 & 1.425 \\
   Swin-Base & 100 & 79.2 & 97.4 & 90 & 85.2 & 98.1 & 93.5 & 63.3 & 1.204 \\
   \bottomrule
\end{tabular}}
\end{table}

\noindent\textbf{The effect of the self-attention module before prediction heads}. Since each RoI feature exclusively covers a single building, it lacks access to the global information of the entire image. Global information becomes instrumental in avoiding the overlap of building edges or points when extracting multiple adjacent buildings. It also proves crucial when extracting polygonal outlines for large buildings. To mitigate this issue, we introduced a self-attention module before feeding the object features into the prediction heads to augment the model's receptive field. Results from experiments using an additional self-attention module and those without are reported in \textcolor{red}{Table \ref{tab:4}}. Both configurations utilize ResNet-50 as the backbone and follow the same training strategy, encompassing 100 training epochs with a learning rate drop at the $54^{\rm th}$ epoch.
\begin{table}
\centering
\caption{The effect of the self-attention (SA) module before prediciotn heads.}
\label{tab:4}
\resizebox{\linewidth}{!}{
\begin{tabular}{lllllllllllll}
   \toprule
   SA & $\uparrow \rm AP$ & $\uparrow \rm AP_{50}$ & $\uparrow \rm AP_{75}$ & $\uparrow \rm AP_{S}$ & $\uparrow \rm AP_{M}$ & $\uparrow \rm AP_{L}$\\
   \midrule
    & 70.5 & 90.9 & 81.4 & 61.9 & 81.1 & 17.5\\
    $\checkmark$ & \textbf{71.1(+0.6)} & \textbf{93.8(+2.9)} & \textbf{82.9(+1.5)} & \textbf{62.8(+0.9)} & \textbf{82.2(+1.1)} & \textbf{32.4(+14.9)}\\
   \bottomrule
\end{tabular}}
\end{table}

To investigate whether the additional self-attention module improves the model's capability to extract outlines of large buildings, we report results using additional metrics, namely $\rm AP_S$, $\rm AP_M$ and $\rm AP_L$. These metrics denote the Average Precision (AP) of small buildings (less than $32^2$ pixels), medium buildings (greater than $32^2$ pixels and less than $96^2$ pixels) and large buildings (greater than $96^2$ pixels), respectively. As can be seen in Table \textcolor{red}{\ref{tab:4}}, it is evident that the incorporation of the additional self-attention module yields significant improvements across various metrics, notably in $\rm AP_{50}$ and $\rm AP_L$. This observation suggests an improvement in the model's instance-level detection of buildings and an improved capacity for extracting outlines of large buildings, which is in alignment with our earlier hypotheses. Moreover, we observe that the incorporation of the self-attention module in PolyR-CNN leads to a significant improvement in $\rm AP_L$ without a corresponding decline in $\rm AP_S$. On the contrary, $\rm AP_S$ experiences a notable enhancement. We speculate that this phenomenon arises from the fact that the \textcolor{red}{guided} ROI features already encapsulate sufficient local information, and the introduced self-attention module merely augments them with additional global information, similar to a supplementary refinement. This contrasts transformer structures where self-attention modules are critical in feature extraction and reasoning.

\noindent\textbf{The effectiveness of the vertex proposal feature}. Given that RoI features provide only a coarse representation of buildings, we introduced the vertex proposal feature to guide the RoI feature to emphasize the building outlines. To evaluate the effectiveness of the vertex proposal feature, we replaced it with the proposal feature proposed by SparseR-CNN \cite{sun2021sparse} and made a comparison. The proposal feature is merely a learnable latent vector without prior knowledge, which increases the difficulty for the model to learn specific geometric information about buildings. In comparison, the vertex proposal feature is extracted from the predicted polygon vertices through a feed-forward neural network, allowing for more effective learning of building instance characteristics. Consequently, it is expected to provide more effective guidance to the RoI feature, emphasizing the geometric outlines of the building polygon. As can be seen in \textcolor{red}{Table \ref{tab:5}}, the utilization of the vertex proposal feature leads to an improvement in all the metrics, especially in terms of $\rm AP_{75}$. This suggests that the vertex proposal feature enables better attention to detailed information of buildings.

\begin{table}
\centering
\caption{The effect of vertex proposal feature.}
\label{tab:5}
\resizebox{\linewidth}{!}{
\begin{tabular}{lllllllllllllll}
   \toprule
    & Backbone & Epoch & $\uparrow \rm AP$ & $\uparrow \rm AP_{50}$ & $\uparrow \rm AP_{75}$ & $\uparrow \rm AP_{boundary}$ & $\rm \downarrow PoLiS$ \\
   \midrule
   Proposal feature \cite{sun2021sparse} & ResNet-50 & 100 & 68.9 & 93.1 & 79.5  & 47.3 & 1.584 \\
   Vertex proposal feature & \textcolor{red}{ResNet-50} & 100 & \textbf{71.1(+2.2)} & \textbf{93.8(+0.7)} & \textbf{82.9(+3.4)} & \textbf{50.0(+2.7)} & \textbf{1.570(-0.014)}\\
   \bottomrule
\end{tabular}}
\end{table}

\begin{figure}[!htbp]
\subfigure{
\begin{minipage}[t]
{0.12\linewidth}
\centering
\includegraphics[width=\linewidth]{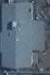}
\end{minipage}
}
\hspace{-5mm}
\subfigure{
\begin{minipage}[t]{0.12\linewidth}
\centering
\includegraphics[width=\linewidth]{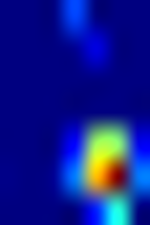}
\end{minipage}
}
\hspace{-5mm}
\subfigure{
\begin{minipage}[t]{0.12\linewidth}
\centering
\includegraphics[width=\linewidth]{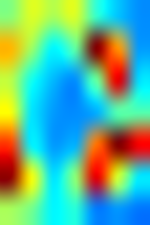}
\end{minipage}
}
\hspace{-5mm}
\subfigure{
\begin{minipage}[t]{0.12\linewidth}
\centering
\includegraphics[width=\linewidth]{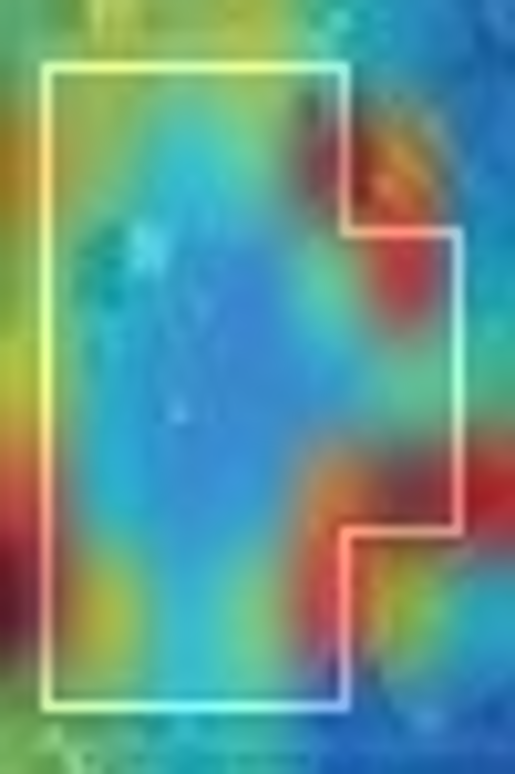}
\end{minipage}
}
\hspace{-5mm}
\subfigure{
\begin{minipage}[t]{0.12\linewidth}
\centering
\includegraphics[width=\linewidth]{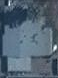}
\end{minipage}
}
\hspace{-5mm}
\subfigure{
\begin{minipage}[t]{0.12\linewidth}
\centering
\includegraphics[width=\linewidth]{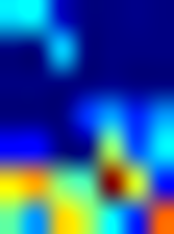}
\end{minipage}
}
\hspace{-5mm}
\subfigure{
\begin{minipage}[t]{0.12\linewidth}
\centering
\includegraphics[width=\linewidth]{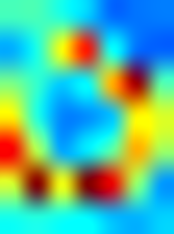}
\end{minipage}
}
\hspace{-5mm}
\subfigure{
\begin{minipage}[t]{0.12\linewidth}
\centering
\includegraphics[width=\linewidth]{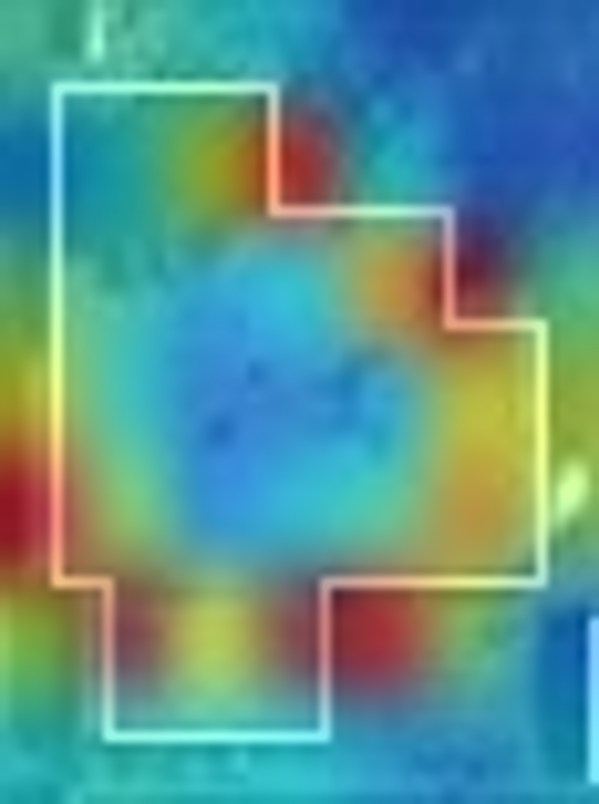}
\end{minipage}
}

\vspace{-3mm}

\subfigure{
\begin{minipage}[t]{0.12\linewidth}
\centering
\includegraphics[width=\linewidth]{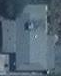}
\end{minipage}
}
\hspace{-5mm}
\subfigure{
\begin{minipage}[t]{0.12\linewidth}
\centering
\includegraphics[width=\linewidth]{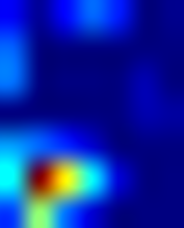}
\end{minipage}
}
\hspace{-5mm}
\subfigure{
\begin{minipage}[t]{0.12\linewidth}
\centering
\includegraphics[width=\linewidth]{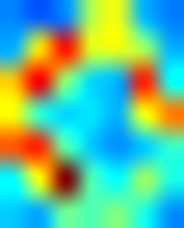}
\end{minipage}
}
\hspace{-5mm}
\subfigure{
\begin{minipage}[t]{0.12\linewidth}
\centering
\includegraphics[width=\linewidth]{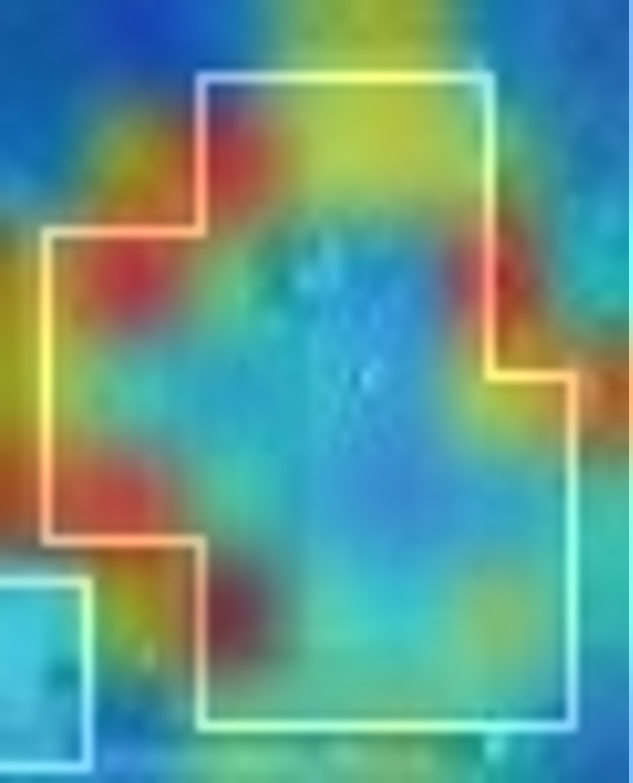}
\end{minipage}
}
\hspace{-5mm}
\subfigure{
\begin{minipage}[t]{0.12\linewidth}
\centering
\includegraphics[width=1\linewidth]{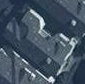}
\end{minipage}
}
\hspace{-5mm}
\subfigure{
\begin{minipage}[t]{0.12\linewidth}
\centering
\includegraphics[width=1\linewidth]{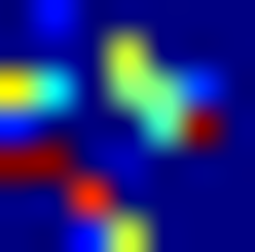}
\end{minipage}
}
\hspace{-5mm}
\subfigure{
\begin{minipage}[t]{0.12\linewidth}
\centering
\includegraphics[width=1\linewidth]{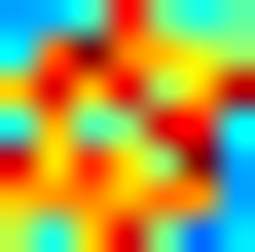}
\end{minipage}
}
\hspace{-5mm}
\subfigure{
\begin{minipage}[t]{0.12\linewidth}
\centering
\includegraphics[width=1\linewidth]{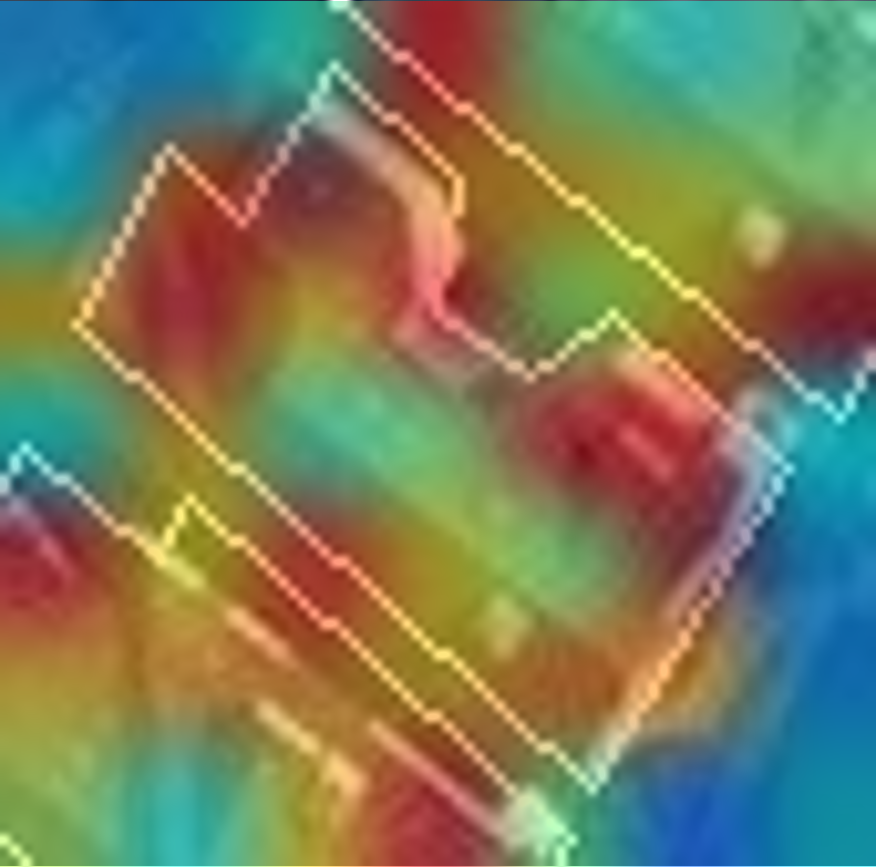}
\end{minipage}
}

\vspace{-3mm}

\subfigure{
\begin{minipage}[t]{0.12\linewidth}
\centering
\includegraphics[width=\linewidth]{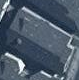}
\end{minipage}
}
\hspace{-5mm}
\subfigure{
\begin{minipage}[t]{0.12\linewidth}
\centering
\includegraphics[width=\linewidth]{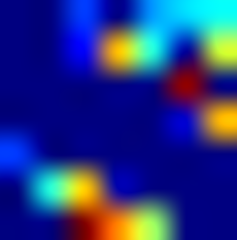}
\end{minipage}
}
\hspace{-5mm}
\subfigure{
\begin{minipage}[t]{0.12\linewidth}
\centering
\includegraphics[width=\linewidth]{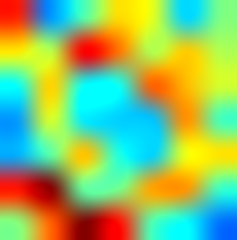}
\end{minipage}
}
\hspace{-5mm}
\subfigure{
\begin{minipage}[t]{0.12\linewidth}
\centering
\includegraphics[width=\linewidth]{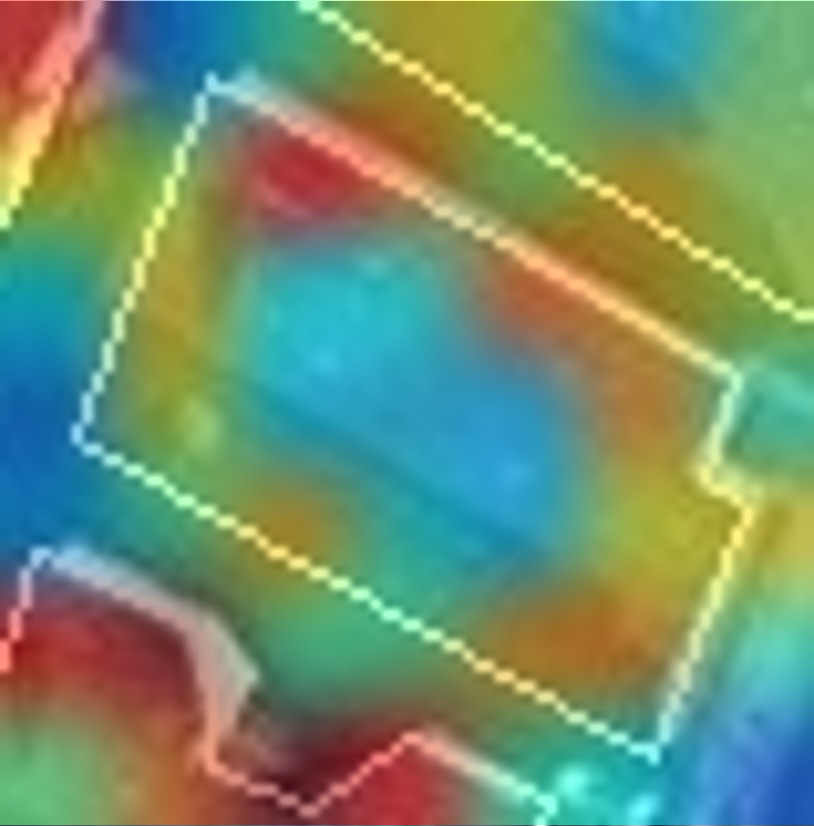}
\end{minipage}
}
\hspace{-5mm}
\subfigure{
\begin{minipage}[t]{0.12\linewidth}
\centering
\includegraphics[width=1\linewidth]{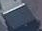}
\end{minipage}
}
\hspace{-5mm}
\subfigure{
\begin{minipage}[t]{0.12\linewidth}
\centering
\includegraphics[width=1\linewidth]{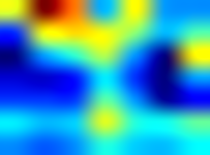}
\end{minipage}
}
\hspace{-5mm}
\subfigure{
\begin{minipage}[t]{0.12\linewidth}
\centering
\includegraphics[width=1\linewidth]{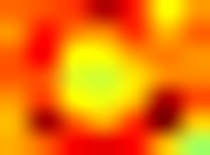}
\end{minipage}
}
\hspace{-5mm}
\subfigure{
\begin{minipage}[t]{0.12\linewidth}
\centering
\includegraphics[width=1\linewidth]{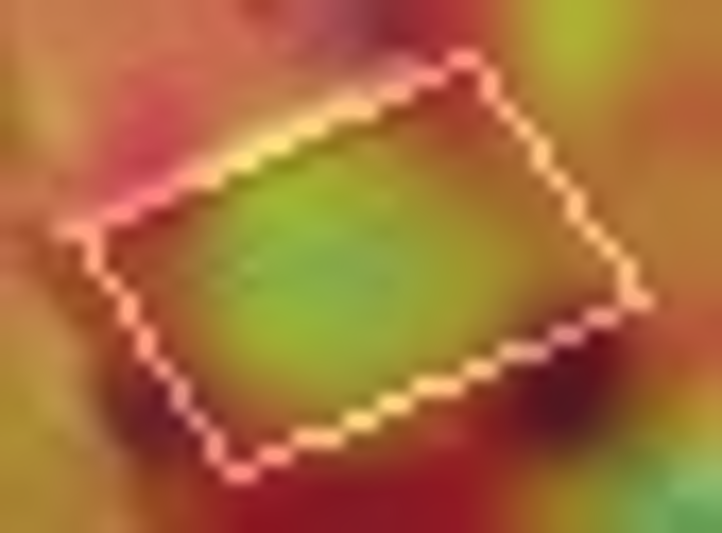}
\end{minipage}
}
\caption{Illustrating the bounding boxes (the first column), the original RoI features (the second column), the RoI features \textcolor{red}{guided} by the vertex proposal features (the third column), and the mapping of \textcolor{red}{guided} RoI features on the reference building polygons (the fourth column). Columns five to eight are arranged in a similar manner. The redder the area on the figure, the higher the value; the bluer the area, the lower the value.}
\label{Fig:2}
\end{figure}

To better demonstrate the efficacy of the proposed vertex proposal feature, we visualized the RoI features guided by the vertex proposal feature alongside those without such guidance, as illustrated in \textcolor{red}{Figure \ref{Fig:2}}. The RoI features are upsampled to the size of the bounding boxes for ease of comparison. The regions on the heatmaps that appear red correspond to higher values, while those that appear blue indicate lower values. As seen in \textcolor{red}{Figure \ref{Fig:2}}, the original RoI features exhibit the minimal capacity to capture meaningful building outline information, while the guided RoI features effectively highlight the geometric features of building contours, especially at corners. This precisely aligns with the expected function of the vertex proposal feature.

\section{Limitations and Future work}
\label{Sec:limit}
Although PolyR-CNN has demonstrated excellent performance in both accuracy and computational efficiency, it still has several limitations. Firstly, PolyR-CNN is prone to generate redundant vertices around building corners. \textcolor{red}{Several potential reasons are identified.} 
First of all, the iterative refinement structure only considers building polygons and bounding boxes, while neglects the corner classification scores. Therefore, the object features lack focus on corner classification. 
Besides, PolyR-CNN predicts $M$ vertices for each building polygon. $M$ is set to 96 to cover most buildings. 
This dense distribution of vertices along the building contour poses a challenge for the model to effectively distinguish closely situated vertices.

\textcolor{red}{Secondly, PolyR-CNN tends to generate irregular shapes when handling buildings with complex structures, especially evident in the inner outlines of buildings in the Inria dataset. 
This phenomenon can be attributed to the model treating all inner outlines as the same class. 
The inner outlines often refer to courtyards which include diverse objects such as vegetation, ground and vehicles. 
This diversity confuses the model in identifying these outlines as a single class. 
Moreover, for fine-grained building structures, instance-level interaction may be too coarse for precise point-level detection.}

In the future work, we aim to enhance the corner classification performance by \textcolor{red}{enabling vertex-level interaction and incorporating corner class information during the training}. 
We will also try to extend the model to a multi-class outline extraction framework, enabling the extraction of various objects beyond buildings. 
Last but not least, PolyR-CNN currently adopts only the L1 loss for polygon coordinate regression. However, existing methods have shown the effectiveness of additional geometrical supervision signals and loss terms, such as the attraction field map \cite{xu2022accurate} and the corner angle loss \cite{zorzi2022polyworld}. 
We will also explore the integration of these techniques in our future work. 

\section{Conclusion}
In this paper, we introduced PolyR-CNN, a simple, computational efficient and fully end-to-end method for building outline extraction. PolyR-CNN relies only on the RoI features of buildings to directly predict vectorized building polygons. The RoI features are guided by the vertex proposal feature derived from building corner coordinates to focus on geometrical shape information. Notably, PolyR-CNN achieved the best accuracy among end-to-end methods on the CrowdAI dataset and is performing comparable with state-of-the-art multi-stage methods, while exhibiting a lower model complexity and higher computational efficiency. Additionally, PolyR-CNN is also the first end-to-end method demonstrating the capability of extracting buildings with holes on the Inria dataset. \textcolor{red}{As the first R-CNN-based model to directly predict building polygons as ordered vertex sequences,} we hope our work will provide new insights for future research on building outline extraction.

\section{Declaration of competing interest}
The authors declare that they have no known competing financial interests or personal relationships that could have appeared to influence the work reported in this paper.

\section{Acknowledgement}
This publication is part of the project "Learning from old maps to create new ones", with project number 19206 of the Open Technology Programme which is financed by the Dutch Research Council (NWO).


 \bibliographystyle{elsarticle-num} 
 \bibliography{cas-refs}





\end{document}